\documentclass[letter]{llncs}
\usepackage{makeidx}

\usepackage{graphicx} 
\usepackage{amsmath} 
\usepackage{amssymb}  
\makeatletter
\newcommand{\@chapapp}{\relax}%
\makeatother
\usepackage[title,toc,titletoc,header]{appendix}
\usepackage{verbatim}
\usepackage{color}
\usepackage{tabularx}
\usepackage{tabulary}
\usepackage{algorithm}
\usepackage[noend]{algpseudocode}
\usepackage[caption=false]{subfig}
\usepackage{nicefrac}

\usepackage[table,xcdraw]{xcolor}

\title{\LARGE \bf Planning and Resilient Execution of Policies For Manipulation in Contact with Actuation Uncertainty}

\author{Calder Phillips-Grafflin$^{1}$ and Dmitry Berenson$^{2}$%
\institute{$^{1}$Worcester Polytechnic Institute, $^{2}$University of Michigan}
}

\begin{document}

\maketitle
\pagestyle{plain}

\begin{abstract} 
We propose a method for planning motion for robots with actuation uncertainty that incorporates contact with the environment and the compliance of the robot to reliably perform manipulation tasks. Our approach consists of two stages: (1) Generating partial policies using a sampling-based motion planner that uses particle-based models of uncertainty and simulation of contact and compliance; and (2) Resilient execution that updates the planned policies to account for unexpected behavior in execution which may arise from model or environment inaccuracy. We have tested our planner and policy execution in simulated $SE(2)$ and $SE(3)$ environments and Baxter robot. We show that our methods efficiently generate policies to perform manipulation tasks involving significant contact and compare against several simpler methods. Additionally, we show that our policy adaptation is resilient to significant changes during execution; e.g. adding a new obstacle to the environment.
\end{abstract}

\section{Introduction}
\label{sec:intro}
Many real-world tasks are characterized by uncertainty: actuators and sensors may be noisy, and often the robot's environment is poorly modelled. Unlike robots, humans effortlessly perform everyday tasks, like inserting a key into a lock, which require fine manipulation despite limited sensing and imprecise actuation. We observe that humans often perform these tasks by exploiting \emph{contact}, \emph{compliance}, and \emph{resilience}. Using compliance to safely make contact and move while in contact allows us to reduce uncertainty. We also exhibit resilience: when an action fails to produce the desired result, we may withdraw and try again. Seminal motion planning work by Lozano-P{\'e}rez et al. \cite{LMT} shows that incorporating contact and compliance is critical to performing fine motions like inserting a peg into a hole. Building from this work and our observations of human motions, we have developed a motion planner that incorporates contact, compliance, and resilience to generate behavior for robots with actuation uncertainty. 


Motion in the presence of actuation uncertainty is an example of a continuous Markov Decision Process (MDP), adding in sensor uncertainty, the problem becomes a Partially-Observable Markov Decision Process (POMDP). Solving an MDP or POMDP is often framed as the problem of computing an optimal policy $\pi^*$ that maps each state to an action $a$ that maximizes the expected reward (e.g. the probability of reaching the goal). This paper focuses on motion planning with actuation uncertainty, and thus we frame the problem as an MDP. This MDP formulation is representative of the challenges face by low-cost and compliant robots such as Baxter or Raven, which have accurate sensing but noisy actuators.

Instead of planning in the configuration or state-space of the robot, we represent the uncertainty of the state of the robot as a probability distribution over possible configurations, and plan in the space of these distributions---the \emph{belief space}\footnote{The term \textit{belief} is borrowed from POMDP literature, which assumes that the state is partially-observable. Though this paper considers only MDPs, we nevertheless use ``belief'' as it is a convenient and widely-used term for a distribution over states.}. The computational expense of optimal motion planning leads us to adopt a thresholding approach from \emph{conformant} planning \cite{conformant}. Instead of attempting to find a global optimal policy, we seek to generate a \textit{partial} policy that allows a robot with actuation uncertainty to move from start configuration $q_\mathit{start}$ to reach goal $q_\mathit{goal}$ within tolerance $\epsilon_\mathit{goal}$ with at least planning threshold $P_\mathit{goal}$ probability. A partial policy, which maps a subset of possible states to actions rather than a global policy that maps all states to actions, simplifies the problem and is appropriate for the single-query planning problems we seek to solve.

The complexity of robot kinematics and dynamics preclude analytical modeling of compliance and contact for practical, high-dimensional problems, and thus we rely on the ability to forward simulate the state of the robot given a starting state and action. In the presence of uncertainty, individual actions may have multiple distinct outcomes: for example, when trying to insert a key into a lock, some attempts will succeed in inserting the key, while some will miss the keyhole. In advance of performing such an action, we cannot \emph{select} between desired outcomes (as is assumed in \cite{particleRRT}). However, we can \emph{distinguish} between the outcomes after the action is executed. We directly incorporate this behavior into our planner using \emph{splits} and \emph{reversibility}. Splits are single actions that produce multiple distinct outcomes, which we distinguish between using a series of clustering algorithms. Reversibility is the ability of a specific action and outcome to be ``undone'' and return to the previous state, which allows the robot to attempt the action again. Of course, the planner may not accurately model the outcomes of every action, so we incorporate an online adaptation process to update the planned policy during execution to reflect the results of actions.

Our primary contributions are thus 1) incorporating contact and compliance into policy generation, thus allowing contacts that other planners would discard but that, in fact, can be used to reduce uncertainty; and 2) introducing resilience into policy execution and thus significantly increasing the probability of successfully completing the task. Our experiments with simulated test environments suggest that our planner efficiently generates policies to reliably perform motion for robots with actuation uncertainty. We apply our methods to problems in $SE(2)$, $SE(3)$, and a simulated Baxter robot ($\mathbb{R}^7$) and show performance improvements over simpler methods and the ability to recover from an unanticipated blockage.

\section{Related Work}
\label{sec:relatedwork}
Planning motion in the presence of actuation uncertainty dates back to the seminal work of Lozano-P\'{e}rez et al. \cite{LMT}, which introduced pre-image backchaining. A pre-image, i.e. a region of configuration space from which a motion command attains a certain goal recognizably, was used in a planner that produced actions guaranteed to succeed despite pose and action uncertainty. However, constructing such pre-images is prohibitively computationally expensive \cite{finemotioncomputability,backprojections}.

In its general form, belief-space planning is formulated as a Partially-Observable Markov Decision Process (POMDP), which are widely known to be intractable for high-dimensional problems. However, recent developments of general approximate point-based solvers such as SARSOP \cite{SARSOP} and MCVI \cite{MCVI} have made considerable progress in generating policies for complex POMDP problems. For some lower-dimensional robot motion problems like \cite{contactpomdp}, the POMDP can be simplified by extracting the part of the task that incorporates uncertainty (e.g. the position of an item to be grasped) and applying off-the-shelf solvers to the problem. Others have investigated learning approaches \cite{policysearch} for similar problems; however, we are interested in planning because we want our methods to generalize to a broad range of tasks without collecting new training data.



Several sampling-based belief-space planners have been developed \cite{particleRRT,beliefroadmap,RRBT,FIRM,stochasticmotionroadmap}. Others have evaluated the belief-space distance functions \cite{beliefdistance} and show that the selection of distance function greatly impacts the performance of the planner. Additionally, approaches using LQG and LQR controllers \cite{FIRM,LQGMP,incrementastochastic} and trajectory optimizers \cite{copt,sigmahulls} have been proposed. These approaches use Gaussian distributions to model uncertainty, but such a simple distribution cannot accurately represent the belief of a robot moving in contact with obstacles, where belief may lose support in one or more dimensions, or the state may become trans-dimensional. Other approaches like \cite{particleRRT} use a set of particles to model belief like a particle filter; while we also use a particle-based representation, our approach more accurately captures the behavior of splits and also includes resilience during execution.

The importance of compliance has long been known, with \cite{LMT} demonstrating the important role of compliance in performing precise motion tasks. Sampling-based motion planning for compliant robots has been previously explored in \cite{pushingRRT}, albeit limited to disc robots with simplified contact behavior. We draw from these methods, but our approach differs significantly from previous work by incorporating contact and compliance directly into the planning process by using forward simulation like the kinodynamic RRT \cite{kinodynamicRRT}. A major advantage over existing methods is that the policies we generate are not fixed; instead, we update them online during execution, which allows us to reduce the impact of differences between our planning models and real-world execution conditions.

\section{Problem Statement}
\label{sec:problemstatement}
We consider the problem of planning motion for a \emph{controlled} \emph{compliant} robot $R$ with configuration space $\mathcal{Q}$ in an environment with obstacles $E$. For given start ($q_\mathit{start}$) and goal ($q_\mathit{goal}$), we seek to produce motion which allows the robot to reach $q_\mathit{goal}$ within tolerance $\epsilon_\mathit{goal}$ with at least $P_\mathit{goal}$ probability.


The robot is assumed to have actuation uncertainty modelled by $q_\mathit{t+1} = q_\mathit{t} + (\Delta q_\mathit{t} + r_\mathit{\Delta q})$ in which the next configuration $q_\mathit{t+1}$ is the result of the previous configuration $q_\mathit{t}$, control input $\Delta q$ and actuation error $r_\mathit{\Delta q}$. We assume that a function $\bold{F}$, which models the probability distribution of the uncertainty, is available from which to sample $r_\mathit{\Delta q} \sim \bold{F}(\Delta q)$ for a given $\Delta q$.  Due to this actuation uncertainty, when executing actions in our planner the result is a belief distribution $b$. The robot is compliant, meaning that for a motion from collision-free $q_\mathit{current}$ to colliding $q_\mathit{desired}$, the resulting configuration $q_\mathit{result}$ will be in contact and the robot will not damage itself or the environment.

Since the motion of the robot is uncertain, a path $\tau$ that is a discrete sequence of configurations may not be robust to errors. Instead, we wish to produce a partial policy $\pi : \mathcal{Q}' \rightarrow A$ that maps $\mathcal{Q}' \subseteq \mathcal{Q}$ to actions $A$ such that for a configuration $q \in \mathcal{Q}'$, the policy returns an action to perform. Even $\pi$ may not always be robust to unexpected errors, therefore during execution we wish to detect actions that do not reach their expected results; i.e. when an action produces $q_\mathit{result} \notin \mathcal{Q}'$. In such an event, we wish to adapt $\mathcal{Q}'$ and $\pi$ such that $q_\mathit{result} \in \mathcal{Q}'$ and continue attempting to complete the task.

\section{Methods}
\label{sec:methods}

We have developed a motion planner consisting of an anytime RRT-based global planner and a local planner that uses a kinematic simulator to model robot behavior. Together, they produce a set of solution paths $S$, where each solution $s \in S$ is a sequence of nodes $n_i = (b_i, a_i)$, in which $b_i$ is the belief distribution for $n_i$ and $a_i$ is the action that produced $b_i$. Using this set of solution paths, we construct a single partial policy $\pi$. As $\pi$ is queried during execution, we update the policy to reflect the ``true'' state observed during the execution process.

\begin{figure*}[t]
\centering
\label{fig:expansion}
\includegraphics[width=\textwidth]{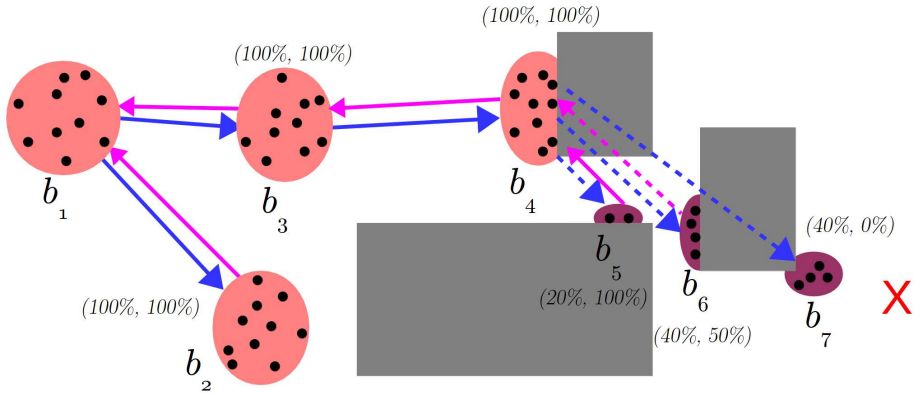}
\caption{\small{Our belief-space RRT extending toward a random target (red X) from $b_\mathit{4}$. Due to compliance, the particles (dots) can slide along the obstacles (gray). Solid blue edges denote 100\% probability edges, dashed edges denote a split resulting in multiple states; solid magenta edges denote 100\% reversible edges, while dashed edges denote lower reversibility. Because the extension is attempting to move through a narrow passage, particles separate and a split occurs, resulting in three distinct states ($b_\mathit{5}, b_\mathit{6}, b_\mathit{7}$).}}
\label{fig:particleexamples}
\end{figure*}

Because it is difficult to model the belief state in contact using a parametric distribution, we use a particle-based approach similar to \cite{particleRRT} in which we represent the belief $b_i$ of node $n_i$ with a set of configurations $q_1, q_2,...,q_n$ that are forward-simulated by the local planner. Like previous work \cite{particleRRT}, we expect that performing some actions will result in multiple qualitatively different states as illustrated in Figure \ref{fig:particleexamples} (e.g. in contact with an obstacle some particles will become stuck on the obstacle while others slide along the surface). These distinct parts of the belief state, which we refer to as \emph{splits}, are distinguished in our planner by a series of clustering operations. To ensure that all actions are adequately modeled, a fixed number of particles $N_\mathit{particles}$ is used to simulate every action; since splits reduce the number of particles at a given state, a new set of particles must be resampled for these states to avoid particle starvation.

It is important to understand that we cannot select between the different result states of a split when performing the action; however, we can \emph{distinguish} using our clustering methods if we have reached an undesirable result. To be \emph{resilient} to such errors, we incorporate the ability to reverse the action back to the previous state and try the action again. Clearly, not all actions will be reversible, so we perform additional simulation to estimate the ability to reverse each action after identifying the resulting states.

We first introduce our RRT-based global planner that uses a simulation-based local planner and a series of particle clustering methods to generate policies incorporating actuation uncertainty, and then discuss our online policy execution and adaptation that enables resiliency to unexpected behavior encountered during execution.


\subsection{Global planner}
\label{sec:globalplanner}

Until it reaches time limit $t_\mathit{planning}$, our global planner iteratively grows a tree $T$ using the local planner to extend the tree towards a sampled configuration $q_\mathit{target}$. Like the RRT, $q_\mathit{target}$ is either a uniformly sampled $q_\mathit{rand} \in \mathcal{C}$, or with some probability, the goal $q_\mathit{goal}$. Each time we sample a $q_\mathit{target}$, we select the closest node in the tree $n_\mathit{near} = \text{argmin}_\mathit{n_i \in T} \Call{Proximity}{b_i, q_\mathit{target}}$. The local planner plans from $n_\mathit{near}$ towards $q_\mathit{target}$ and returns new nodes $\mathcal{N}_\mathit{new}$ and edges $\mathcal{E}_\mathit{new}$ that grow the tree. We check each new node $n_\mathit{new} \in \mathcal{N}_\mathit{new}$ to see if it meets the goal conditions, and if so, add the new solution path to $S$.

We also incorporate several features distinct from the RRT. First, using $\Call{Proximity}{}$ we consider more than distance when selecting the nearest neighbor node $n_\mathit{near}$. We want to bias the growth of the tree toward nodes that can be reached with higher probability and have more concentrated $b_i$, so we incorporate weighting using $P(n_\mathit{start} \rightarrow n_i)$, the probability the entire path from $n_\mathit{start}$ to $n_i$ succeeds, and $\mathrm{var}(n_i)$, the variance of $b_i$. The proximity of a node $n_i$ to a configuration $q$ is given by the following equation:

\begin{multline}
\label{eq:proximity}
\Call{Proximity}{n_i, q} = \mathrm{dist}(\mathrm{expect}(b_i), q) \\ * [ (1 - P(n_\mathit{start} \rightarrow n_i)) * \alpha_\mathit{P} + (1 - \alpha_\mathit{P}) ] [ \mathrm{erf}(|\mathrm{var}(b_i)|_1) * \alpha_\mathit{V} + (1 - \alpha_\mathit{V}) ]
\end{multline}

Here, $\mathrm{expect}(b_i) = q_\mathit{expected}$ is the expected value of the belief distribution $b_i$ and $\mathrm{dist}(q_\mathit{expected}, q)$ is the $\mathcal{C}$-space distance function. Two weights $\alpha_\mathit{P}$ and $\alpha_\mathit{V}$ control the effect of the probability and variance weighting, respectively. Values of $\alpha_\mathit{P}$ and $\alpha_\mathit{V}$ closer to 1 increase the effect of the weighting, while values closer to 0 increase the effect of the $\mathcal{C}$-space distance.
Using the error function $\mathrm{erf}(x) = \nicefrac{2}{\sqrt{\pi}} \int_\mathit{0}^{x} e^{-t^{2}} dt$ maps variance in the range $[0,\inf)$ to the range $[0,1)$ to simplify computation. Previous work in belief-space planning has used a range of distance functions, such as L1, Kullback-Leibler divergence, Hausdorff distance, or Earth Mover's Distance (EMD) \cite{beliefdistance}; however, many of these choices only provide useful distances between belief states with overlapping support. While EMD encompasses both the $C$-space distance and probability mass of two beliefs, it is expensive to compute. Since most of our distance computations are between beliefs with non-overlapping support, the $\mathcal{C}$-space distance between expected configurations is an efficient approximation \cite{beliefdistance}. 

Second, we cannot simply test if $n_\mathit{new} = q_\mathit{goal}$, since the $P(n_\mathit{start} \rightarrow n_\mathit{new})$ may be low; instead, we check if a new solution has been found. To be a solution, the probability $n_\mathit{new}$ reaches the goal must be greater than $P_\mathit{goal}$, i.e. the product of $P(n_\mathit{start} \rightarrow n_\mathit{new})$ and $|{q \in b_\mathit{new} | \mathrm{dist}(q, q_\mathit{goal}) \leq \epsilon_\mathit{goal}}|$.
Finally, once a path to the goal has been found, we continue planning to find alternative paths. We want to encourage a diverse range of solutions, so once a solution path has been found, we remove nodes on solution branches from consideration for nearest neighbor lookups. This process recurses towards the root of the planner's tree $T$ until it either reaches the root node $n_\mathit{start}$ or a node $n_\mathit{i}$ which is the result of a split. Once the base of the solution branch is found, we remove the branch from nearest neighbors consideration and continue planning until reaching $t_\mathit{planning}$.

\subsection{Local planner}
\label{sec:localplanner}

Our local planner grows the planner tree $T$ from nearest neighbor node $n_\mathit{near}$ towards a target configuration $q_\mathit{target}$ by forward-propagating belief using $\Call{Extend}{}$ to produce one or more result nodes $n_\mathit{new} \in \mathcal{N}_\mathit{new}$ and edges $e_\mathit{new} \in \mathcal{E}_\mathit{new}$ (recall that splits may occur). To improve the time-to-first-solution, the local planner operates like RRT-Connect, repeatedly calling $\Call{Extend}{}$, until a solution is found, whereupon it switches to RRT-Extend, calling $\Call{Extend}{}$ only once, to improve coverage of the space and encourage solution diversity. Note that the RRT-Connect behavior is stopped if an extension results in a split.

$\Call{Extend}{}$ forward-simulates particles $Q_\mathit{initial}$ from node $n_\mathit{near}$ towards $q_\mathit{target}$, clusters the resulting particles $Q_\mathit{results}$ into new nodes $\mathcal{N}_\mathit{new}$, and computes the transition probabilities. As previously discussed, we simulate every action with the same number of particles. If node $n_\mathit{near}$ is \emph{not} the result of a split, and thus $b_\mathit{near}$ contains a full set of particles, then we simply copy $b_\mathit{near}$ to use for simulation. If, instead, $n_\mathit{near}$ is the result of a split, then we uniformly resample $N_\mathit{particles}$ particles from $b_i$. We then simulate the extension toward $q_\mathit{target}$ for each particle. Any simulation engine that simulates contact and compliance could be used, but the simulation should be as fast as possible to minimize planning time. In our experiments, we used an approximate kinematic simulator described in Appendix \ref{app:kinematicsimulation}. The resulting particles are then grouped into one or more clusters using $\Call{ClusterParticles}{}$, which we describe in Section \ref{sec:particleclustering}. For each cluster $Q_\mathit{cluster}$, we form a new node $n_\mathit{new} = (b_\mathit{new}, a_\mathit{new})$ with belief $b_\mathit{new} = Q_\mathit{cluster}$ and action $a_\mathit{new} = q_\mathit{target}$. In the case of splits, where multiple nodes are formed, we assign $P(n_\mathit{near} \rightarrow n_\mathit{new,i}) = |b_\mathit{new,i}| / N_\mathit{particles}$. We then estimate the probability that action $a_\mathit{new}$ can be reversed from node $n_\mathit{new}$ by simulating $N_\mathit{particles}$ particles back towards node $n_\mathit{near}$. Note that some particles may become stuck while reversing, and thus the probability of reversing the action may not be 1.

The ability to reverse an action allows us to detect an undesired outcome, reverse to the parent node, and retry the action until we either reach the desired outcome or become stuck. Thus, we estimate the \emph{effective} probability $P(n_\mathit{near} \rightarrow n_\mathit{new})_\mathit{effective}$ for each node $n_\mathit{new}$ by estimating the probability that a particle has reached $n_\mathit{new}$ after $N_\mathit{attempt}$ attempts, where at each attempt, particles that have not reached the $n_\mathit{new}$ try to return to $n_\mathit{near}$ and try again. 

\emph{Analysis} -- 
The planner always stores $N_\mathit{stored} = N_\mathit{actions} N_\mathit{particles}$ particles. For every action $N_\mathit{particles}$ particles are forward-simulated, and all of them are stored in $\mathcal{N}_\mathit{new}$. In the worst case, where every action produces $N_\mathit{particles}$ distinct nodes, the number of particles that must be simulated $N_\mathit{simulated} = N_\mathit{nodes} (N_\mathit{particles} + 1)$, as each node itself is the product of one initial simulation and $N_\mathit{particles}$ simulations are required to estimate reverse probability. In practice, as we discuss in Section \ref{sec:se3results}, the space requirements to perform complex tasks are low as most actions produce a small number of nodes, and the time cost can be reduced by simulating particles in parallel.

\subsection{Particle clustering}
\label{sec:particleclustering}

\begin{figure*}[t]
    \centering
    \subfloat[]{\label{fig:rawparticles}\includegraphics[width=0.2\textwidth]{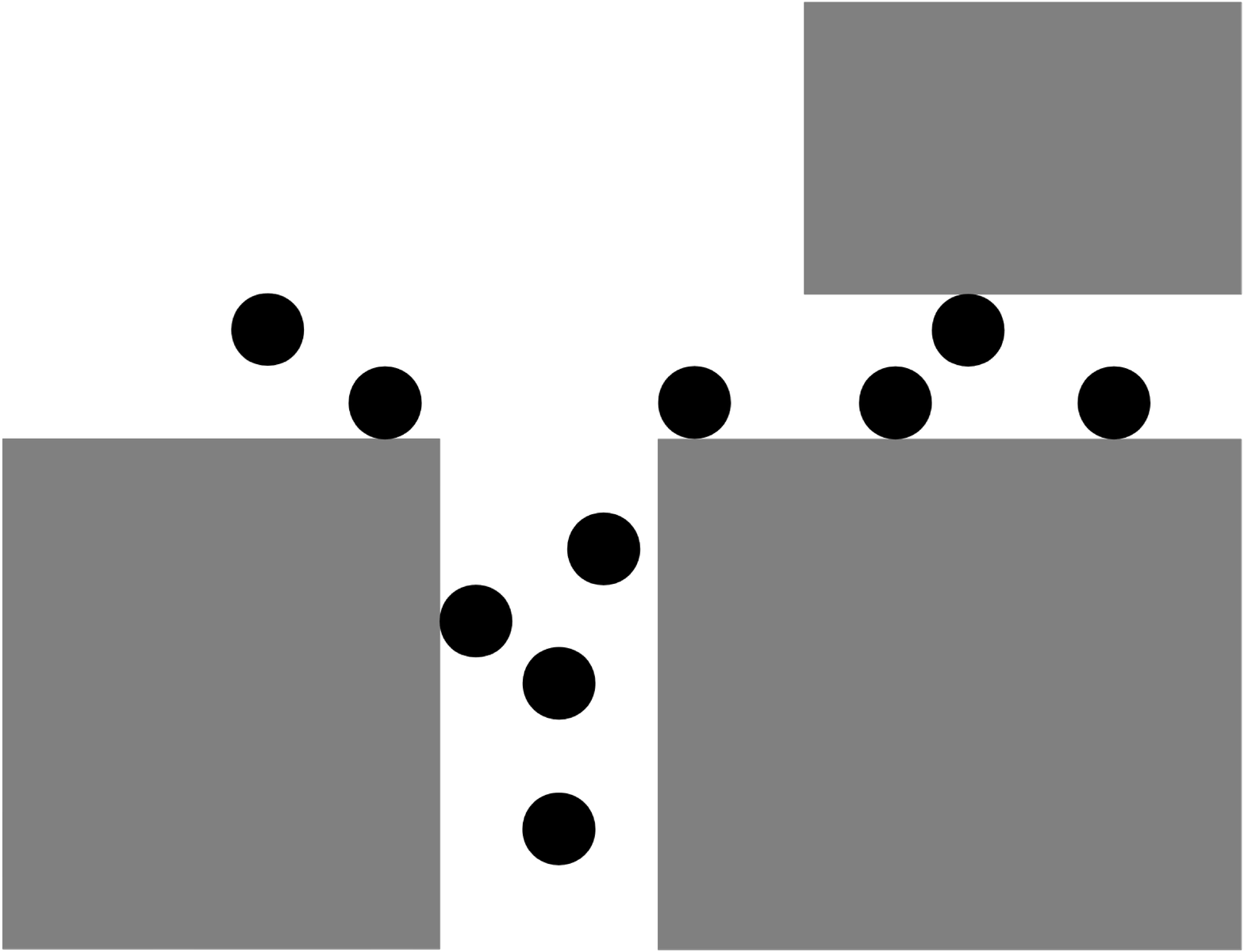}
    }
    \subfloat[]{\label{fig:acclustering}\includegraphics[width=0.2\textwidth]{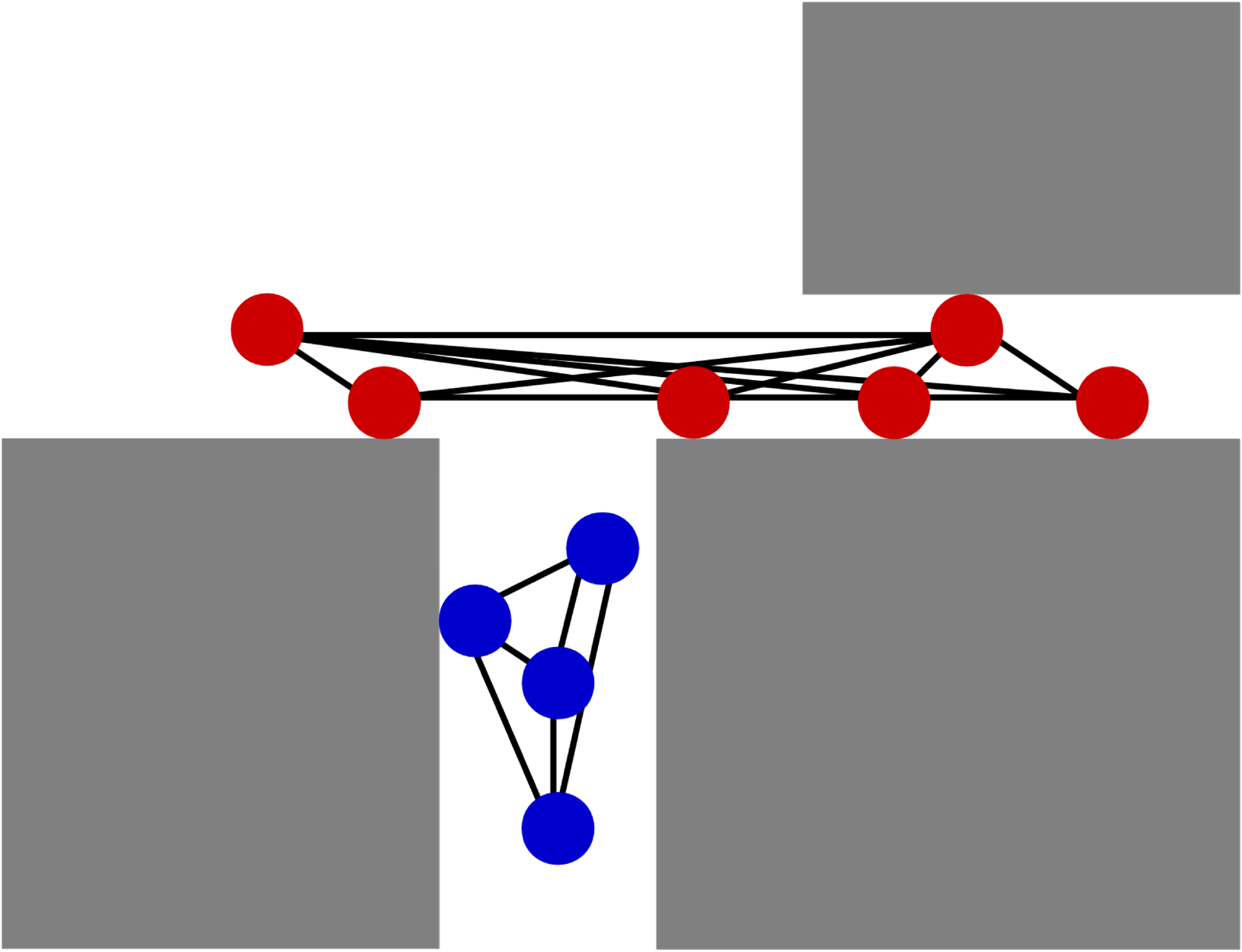}
    }
    \subfloat[]{\label{fig:crsclustering}\includegraphics[width=0.2\textwidth]{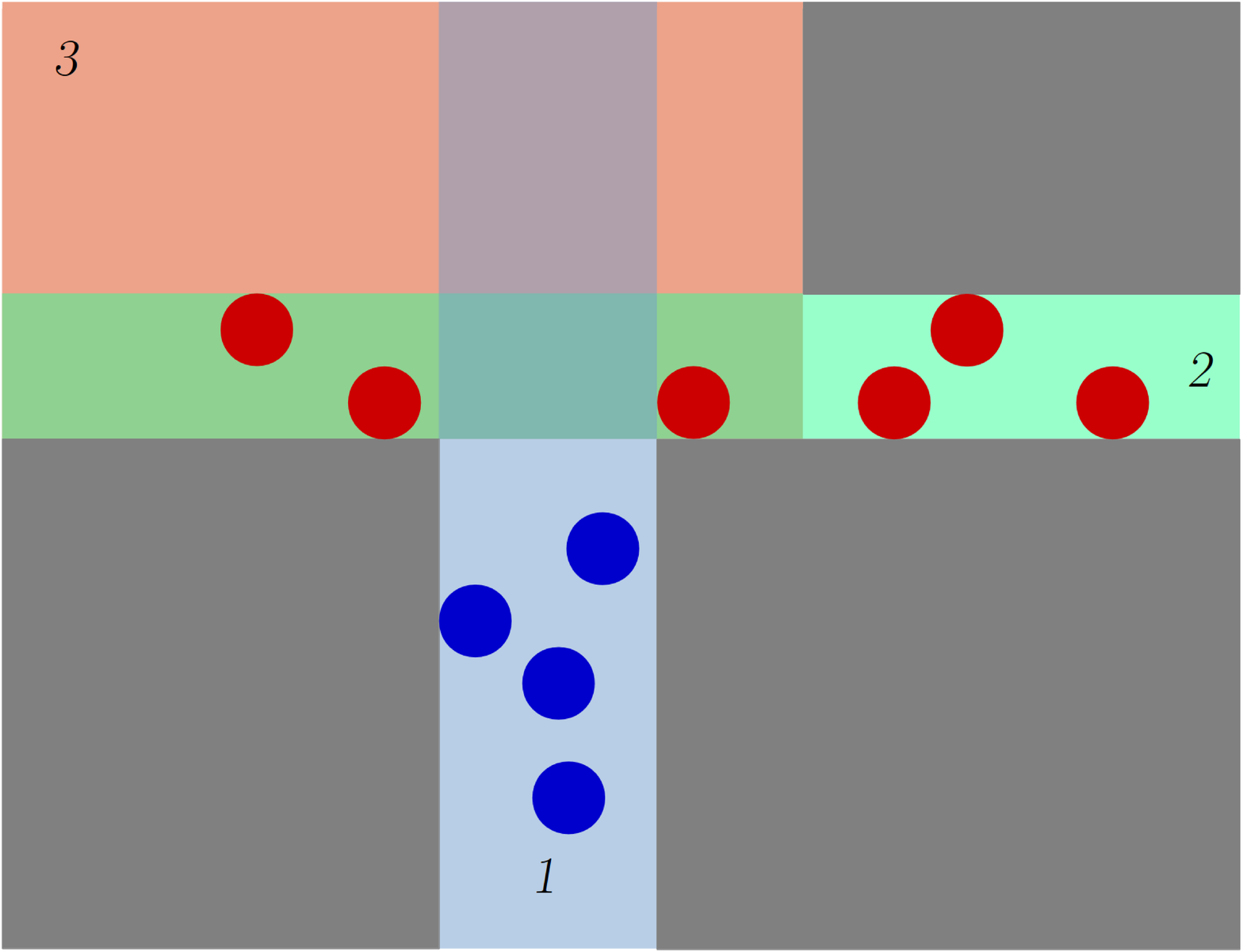}
    }
    \subfloat[]{\label{fig:ptpmclustering}\includegraphics[width=0.2\textwidth]{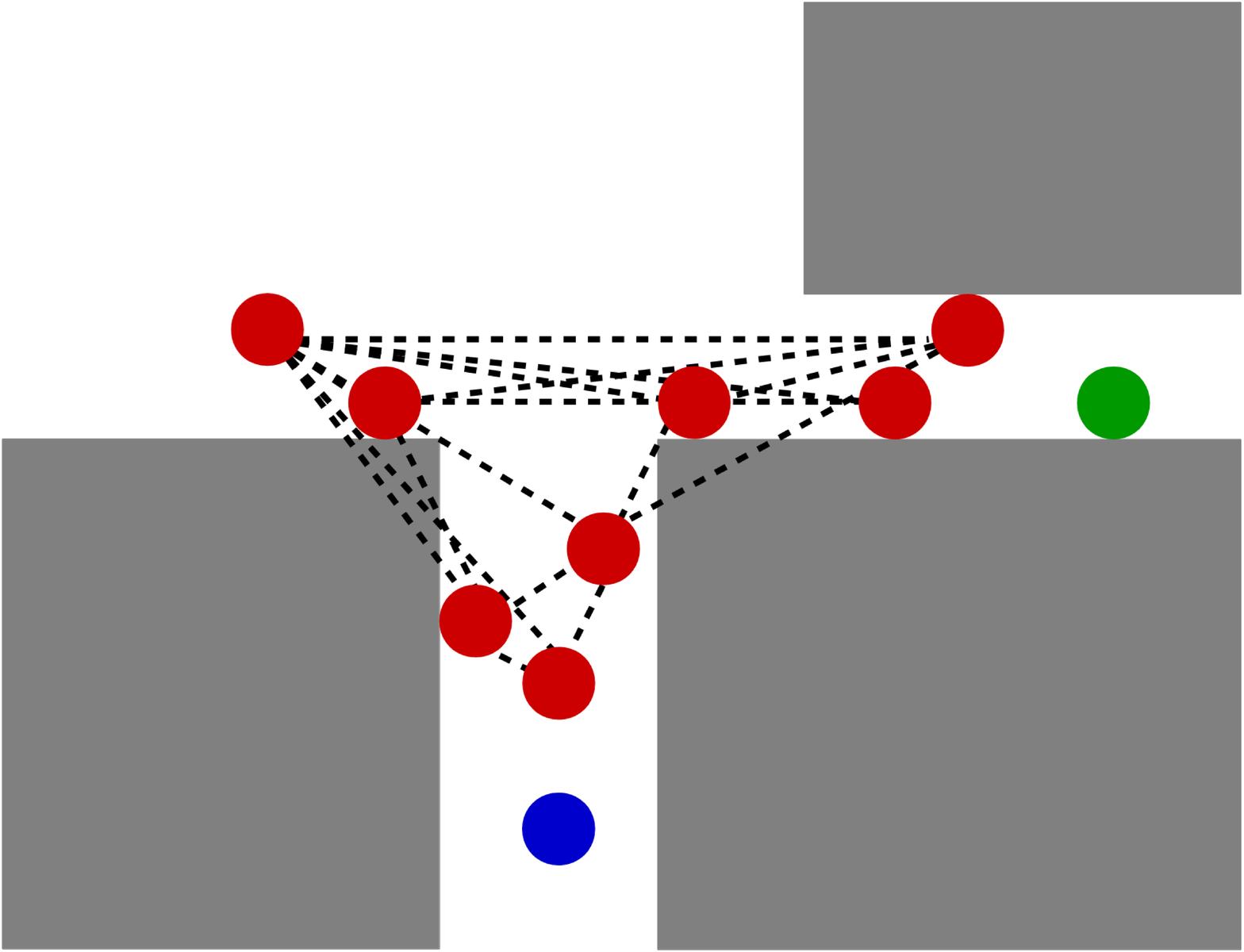}
    }
    \,
    \caption{\small{Our proposed spatial-feature particle clustering methods. (a) The positions of particles after an extension of the planner. (b) Actuation center clustering, with clusters (red, blue) and the straight-line paths for each cluster. (c) Weakly Convex Region Signature clustering, with the three convex regions shown and labeled. (d) Particle movement clustering, with successful particle-to-particle motions shown dashed for the main cluster (red) and two unconnected particles (blue, green).}}
    \label{fig:clusteringmethods}
\end{figure*}

 Intuitively, we want every configuration in a cluster to be reachable from every other one using the local planner. However, testing this directly is computationally expensive, so we also consider two approximate methods. All clustering methods use two successive passes to cluster the configurations resulting from forward simulation: first, a spatial-feature-based pass that groups configurations based on their relationship to different parts of the workspace, and second, a distance-based pass that refines the initial clusters. All of our clustering methods use complete-link hierarchical clustering, as it produces smaller, more dense clusters, while not requiring the number of clusters to be known in advance \cite{numericaltaxonomy,particleRRT}. Below we discuss the ideal approach and our two approximations, shown in Figure \ref{fig:clusteringmethods}. We compare the performance of these methods in Section \ref{sec:se3results}.

\subsubsection{Particle Connectivity (PC) Clustering}
We run the local planner from every configuration to every other configuration and record which simulations reach within $\epsilon_\mathit{goal}$ of the target. For a pair of configurations $q_1, q_2$, going from $q_1$ to $q_2$ may fail while the opposite succeeds; however, to be conservative, we only record success if both executions succeed. We then perform clustering using the complete-link clustering method with distance threshold $0$, where successful simulations correspond to distance $0$ and unsuccessful simulations correspond to distance $1$. Note that this method is very expensive, since it requires simulating $N^2 - N$ particles for $N$ configurations considered.

\subsubsection{Weakly Convex Region Signature (WCR) Clustering}
Intuitively, in many environments a robot can move freely from $q_1$ to $q_2$ if both configurations reside entirely in the same convex region of the workspace. This is also true for some slight concave features, so long as the features do not block the robot. Conversely, for configurations in clearly distinct regions, it is less likely that the robot can move from one configuration to the other.

Illustrated in Figure \ref{fig:crsclustering}, we capture this intuition by recording the position of the robot relative to \emph{weakly convex regions} of the free workspace, to form what we call the \emph{convex region signature}. These regions form a weakly convex covering: individual regions may contain slight concavity, and multiple regions overlap. Techniques such as \cite{weakconvex} exist to automatically compute these regions, but for simple environments these regions can be directly encoded. The convex region signature of a configuration $q$, $WCR(q)$, records the region(s) occupied by every point of the robot at $q$. Distance metric $D_\mathit{WCR}$ between two region signatures $WCR(q_1)$ and $WCR(q_2)$ is the percentage of points in the robot that do \emph{not} share a common region between the signatures. Using this metric, we perform complete-link clustering. We test different thresholds for $D_\mathit{WCR}$ in Section \ref{sec:se3results}. This method allows configurations with some points in a shared region to be clustered together, while separating configurations that share no regions. At runtime, this method requires $N$ computations of $WCR(q)$ and $\nicefrac{(N^2 - N)}{2}$ evaluations of $D_\mathit{WCR}$ to compute all pairwise distances.

\subsubsection{Actuation Center (AC) clustering}
We observe that many successful motions in contact occur when the actuation (or joint) centers of the starting and ending configuration can be connected by collision-free straight lines, so this method checks the straight-line path from the joint centers of one configuration to those of the other configuration. As with the particle movement clustering approach, configurations with successful (collision-free) paths have distance $0$, while those with unsuccessful (colliding) paths have distance $1$. Like the previous approach, clusters are then produced using complete-link clustering with threshold $0$. At runtime, this method requires $\nicefrac{(N^2 - N)}{2}$ checks of the straight-line paths.


\subsection{Partial policy construction}
Once the global planner has produced a set of solution paths $S$, we construct a partial policy $\pi$. Policy construction consists of the following steps:

\begin{enumerate}
    \item{Graph construction} -- An explicit graph is formed, in which the vertices of the graph are nodes $n_i \in S$, and the edges correspond to the edges forming the paths in $S$. An edge $n_i \rightarrow n_\mathit{i + 1}$ is assigned an initial cost $1 / P(n_i \rightarrow n_\mathit{i + 1})$). This means that likely edges receive low cost, which is necessary to compute maximum-probability paths through the graph.
    \item{Edge cost updating} -- The edge costs are updated to reflect the estimated number of attempts needed to successfully traverse the edge by multiplying the cost of the edge by the estimated number of attempts required to reach $P(n_i \rightarrow n_\mathit{i + 1}) \geq P_\mathit{goal}$. This estimate is the complement of the effective probability discussed in Section \ref{sec:localplanner}; instead of computing the probability of reaching a node after a fixed number of attempts, we compute the number of attempts needed to reach the node with $P_\mathit{goal}$ probability. The fewer attempts necessary to traverse the edge, the faster the policy can be executed, and thus this cost represents an expected execution time.
    \item{Dijkstra's search} -- The optimal path from every vertex in the graph to the goal state is computed using Dijkstra's algorithm. This determines the optimal next state (and thus action to perform) for every state in the graph.
\end{enumerate}

\subsection{Partial policy execution and adaptation}
\begin{algorithm}[t!!]
\caption{Partial policy query algorithm}
\label{alg:policyquery}
\begin{algorithmic}
    \Procedure{PolicyQuery}{$S, \pi, q_\mathit{current}, a_\mathit{performed}$}
        \State $\mathcal{N}_\mathit{potential} \gets \left\{ n_i \in S \: |\: a_i = a_\mathit{performed} \right\}$
        \State $\mathcal{N}_\mathit{matching} \gets \left\{ n_i \in \mathcal{N}_\mathit{potential} \: |\: |ClusterParticles(b_i \cup q_\mathit{current})| = 1 \right\}$
        \If {$\mathcal{N}_\mathit{matching} \neq \emptyset$}
            \State $n_\mathit{reached} \gets argmin_\mathit{n_i \in N_\mathit{matching}} \mathrm{DijkstraDistance}(n_i)$
            \State $\Call{IncreaseProbability}{n_\mathit{reached}, a_\mathit{performed}}$;
            \For {$n_i \in \mathcal{N}_\mathit{potential} \: |\: n_i \neq n_\mathit{reached}$}
                \State $\Call{ReduceProbability}{n_i, a_\mathit{performed}}$
            \EndFor
            \State $\pi \gets \Call{ConstructPolicy}{\pi}$
            \If {$P(n_\mathit{reached} \rightarrow q_\mathit{goal}) \geq P_\mathit{goal}$}
                \State $a_\mathit{next} \gets \pi(n_\mathit{reached})$
                \State \textbf{return} $a_\mathit{next}$
            \Else
                \State \textbf{return} failure
            \EndIf
        \Else
            \State $n_\mathit{observed} \gets \{\{q_\mathit{current}\}, a_\mathit{performed}\}$
            \State $S \gets S \cup n_\mathit{observed}$
            \State \textbf{return} $\Call{PolicyQuery}{S, \pi, q_\mathit{current}, a_\mathit{performed}}$
        \EndIf
    \EndProcedure
\end{algorithmic}
\end{algorithm}

At every step during execution, the partial policy $\pi$ is queried for the next action to perform. While we could simply find the ``closest'' node in the policy using a distance function like Equation \ref{eq:proximity}, doing so would discard important information. Not only do we know the configuration $q_\mathit{current}$ that results from executing an action, but we also know the action $a_\mathit{performed}$ we attempted to perform. Using this information, we know exactly which nodes(s) in $\pi$ the robot should have reached. As shown in Algorithm \ref{alg:policyquery}, we first collect all potential result nodes (i.e. those nodes $n_i$ with actions $a_i = a_\mathit{performed}$). We then use our particle clustering method to cluster $q_\mathit{current}$ with the belief $b_i$ of each $n_i$. This clustering tells us if the robot reached a given state (if a single cluster is formed) or not (multiple clusters). In the unlikely (but possible) event that $q_\mathit{current}$ clusters with multiple potential result nodes, we select the ``best'' matching node $n_\mathit{reached}$ using the distance-to-goal computed via Dijkstra's algorithm.

The key contribution of our policy execution is that we adapt the policy $\pi$ to reflect the results of actual execution. If a matching node $n_\mathit{reached}$ is found, we then update $\pi$ to increase the probability that $n_\mathit{reached}$ is the result of $a_\mathit{performed}$. We assign a constant $A_\mathit{importance} \in \mathbb{N}$ that reflects how much we value the results of executing an action compared to the results of simulating a particle during planning. To update the probability, we increase the counts of attempted $N_\mathit{attempts}$ actions and successful $N_\mathit{successful}$ actions, then recompute probability:

\begin{align}
\label{eq:probabilityup}
P(n_\mathit{previous} \rightarrow n_\mathit{reached} | a) = \frac{N_\mathit{successful} + A_\mathit{importance}}{N_\mathit{attempts} + A_\mathit{importance}}
\end{align}

Likewise, we reduce the probability for other potential result states:

\begin{align}
\label{eq:probabilitydown}
P(n_\mathit{previous} \rightarrow n_\mathit{other} | a) = \frac{N_\mathit{successful}}{N_\mathit{attempts} + A_\mathit{importance}}
\end{align}

This update process allows us to learn online, during execution, the true probabilities of reaching states given an action. In effect, the probabilities computed by the global planner serve as an initialization for this online learning. Once updated, we rebuild policy $\pi$ to reflect the new probabilities. If the probability of reaching the goal $P(n_\mathit{reached} \rightarrow q_\mathit{goal})$ is at least $P_\mathit{goal}$, we query $\pi$ for the next action to take. If the probability of reaching the goal has dropped below $P_\mathit{goal}$, policy execution terminates.

However, sometimes no matching node $n_\mathit{reached}$ exists. This means a split occurred during execution that was not captured in $S$ during planning (e.g. an obstacle that is not accurately modelled in $E$, or where the behavior of the simulator diverges from the true robot). To handle this case, we insert a new node $n_\mathit{observed}$ with belief $b_\mathit{observed} = \{q_\mathit{current}\}$ into $S$, and then retry the policy query (which will now have an exactly matching state). To incorporate reversibility, we initially assign new nodes a reverse probability $N_\mathit{attempts}$ = $N_\mathit{successful} = 1$. Thus, the next action selected by the policy will be to return to the previous node. Together with updating probabilities by inserting new states in this manner, we can thus extend the policy to reflect behavior observed during execution that was not captured during the planning process.

\emph{Analysis} -- In the worst case, a policy $\pi$ cannot be executed successfully, and performing every action $a$ results in a new node $n_\mathit{observed}$. For any $A_\mathit{importance} \in \mathbb{N}, P_\mathit{goal} > 0$, adapting the policy will detect failure and terminate in this case.

\emph{Proof} -- For every action $a_\mathit{i+1},...$, node $n_\mathit{observed}$ will be created with a reverse prior $P(n_\mathit{observed} \rightarrow n_\mathit{previous}) = \nicefrac{1}{1}$. If reversing to $n_\mathit{previous}$ fails, we update $P(n_\mathit{observed} \rightarrow n_\mathit{previous}) = \nicefrac{1}{1 + A_\mathit{importance}}$. For the $i$th successive failed reverse and $n_\mathit{observed,i}$ generated, $P(n_\mathit{observed,i} \rightarrow n_\mathit{previous}) = \mathrm{\Pi}_\mathit{i} \frac{1}{1 + A_\mathit{importance}}$. As the number of failed actions increases $P(n_\mathit{observed,i} \rightarrow n_\mathit{previous}) \rightarrow 0$, and thus $P(n_\mathit{observed,i} \rightarrow q_\mathit{goal}) \leq P(n_\mathit{observed,i} \rightarrow n_\mathit{previous}) \rightarrow 0$. Thus eventually $P(n_\mathit{observed,i} \rightarrow q_\mathit{goal})$ will fall below $P_\mathit{goal} > 0$ and execution will terminate. $\Box$

\section{RESULTS}
\label{sec:results}
We present results of testing our planner in simulated $SE(2)$ and $SE(3)$ environments and a simulated Baxter robot. For dynamic simulation during execution, we use the Gazebo simulator. As our kinematic simulator does not consider friction, we use \emph{contact motion controllers} to reduce contact forces (see Appendix \ref{app:dynamicsimulation}). We present statistical results over a range of actuation uncertainty and clustering methods and show that our planner produces policies that allow execution of tasks incorporating contact and robot compliance. We also present statistical results showing that our online policy updating adapts to unexpected behavior during execution. All planning and simulation testing was performed using 2.4 GHz Xeon E5-2673v3 processors. Likewise, all planning was performed with $\Call{Proximity}{}$ weights $\alpha_\mathit{P} = \alpha_\mathit{V} = 0.75$ (see Equation \ref{eq:proximity}), $N_\mathit{attempt}=50$ attempted reverse/repeats of each action, and planning threshold $P_\mathit{goal}=0.51$, such that solutions must be more likely than not to reach the goal.

\subsection{$SE(3)$ simulation}
\label{sec:se3results}

\begin{figure*}[ht!!]
\centering
\subfloat[]{\label{fig:peginholetask}
\includegraphics[width=0.18\textwidth]{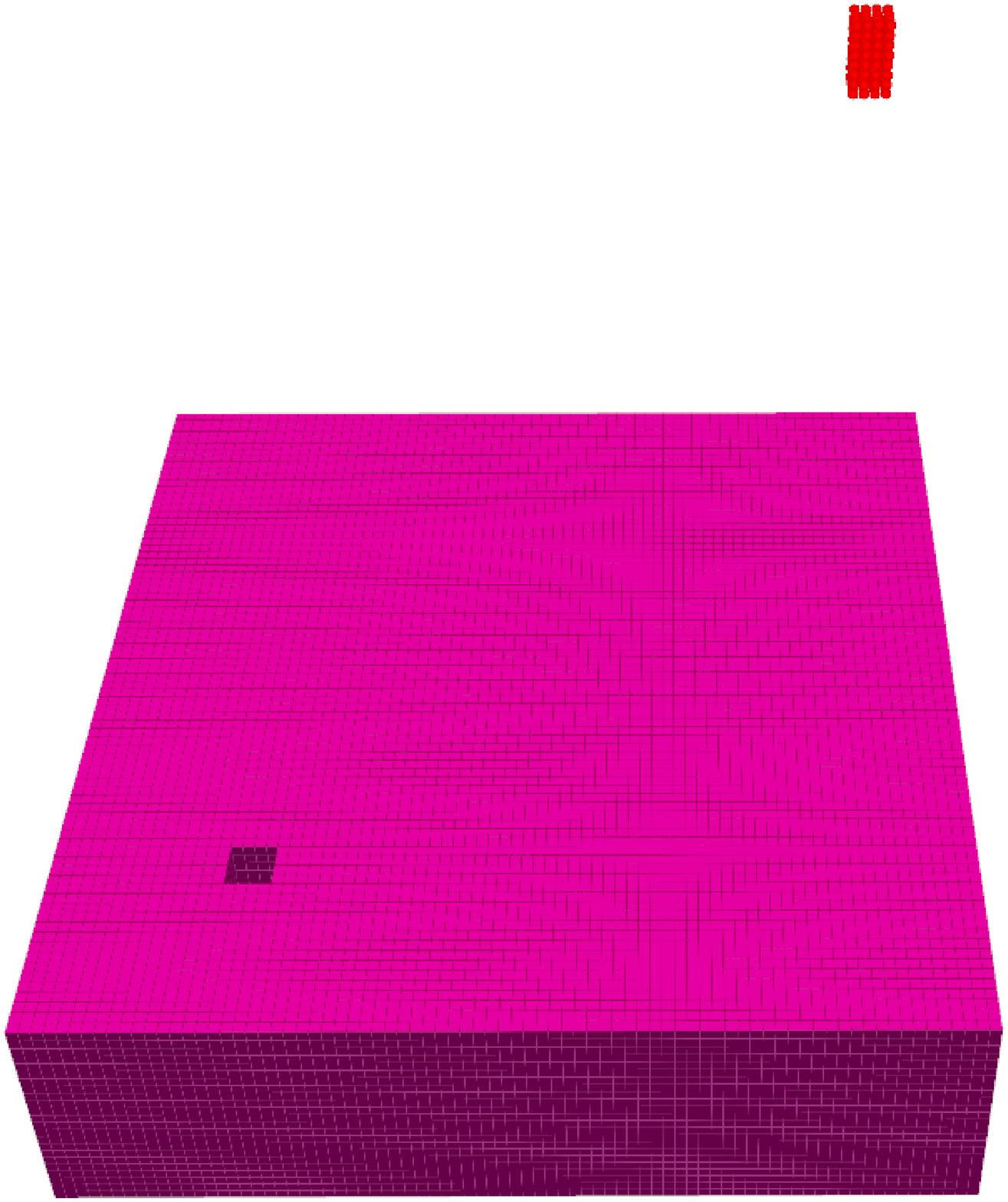}
}
\subfloat[]{\label{fig:examplepolicy}
\includegraphics[width=0.25\textwidth]{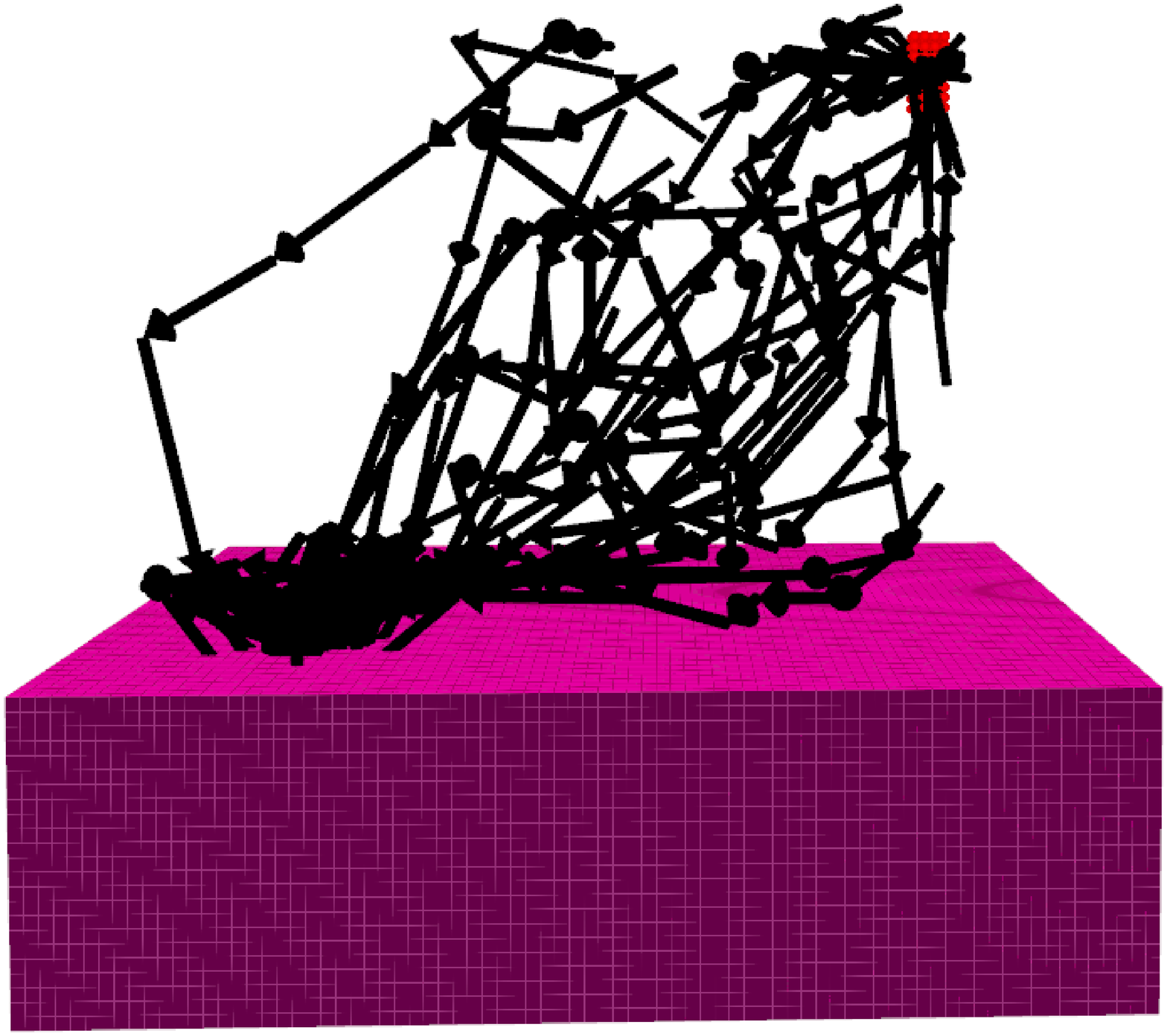}
}
\subfloat[]{\label{fig:examplepolicyexec1}
\includegraphics[width=0.27\textwidth]{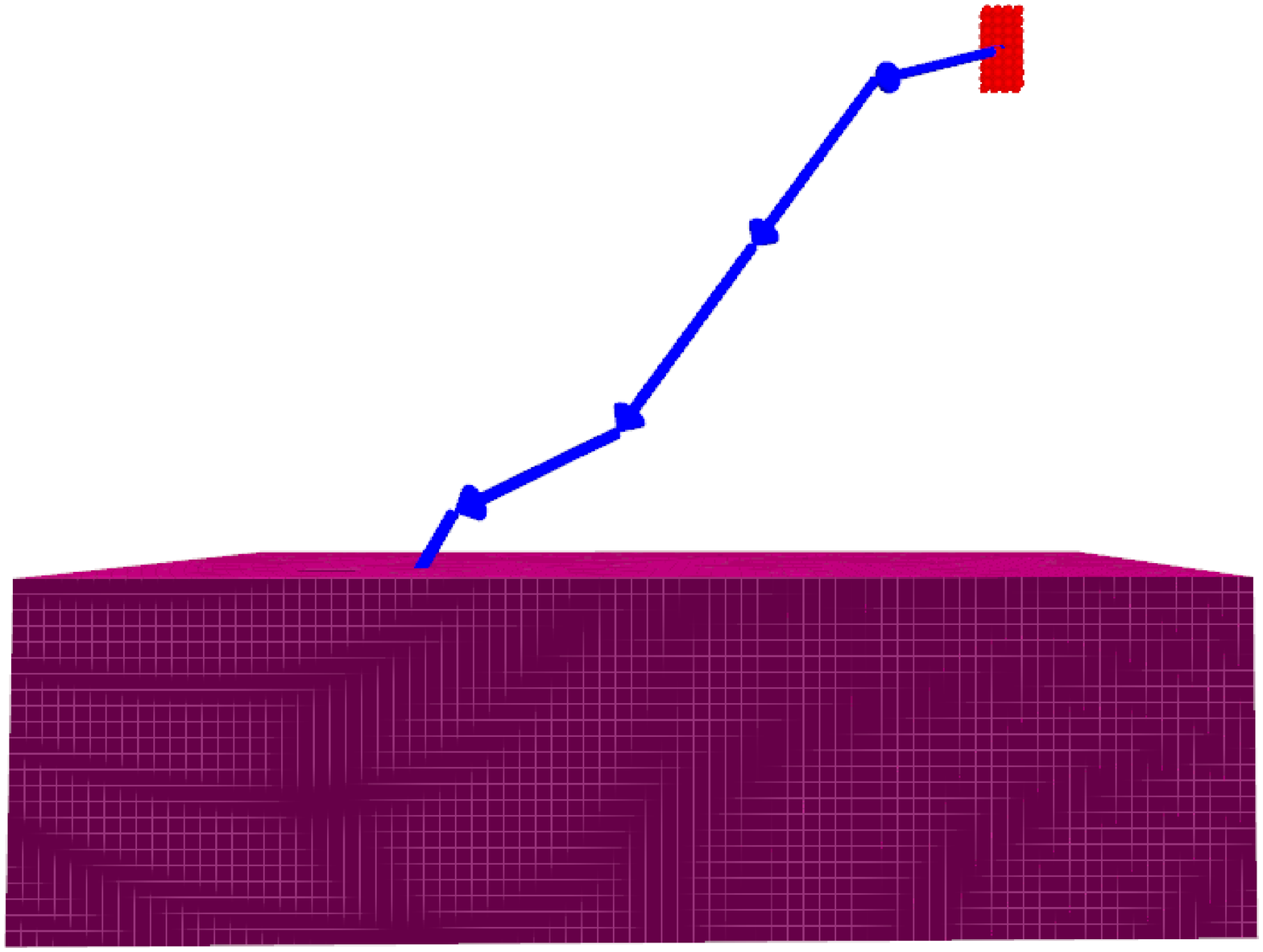}
}
\subfloat[]{\label{fig:examplepolicyexec2}
\includegraphics[width=0.27\textwidth]{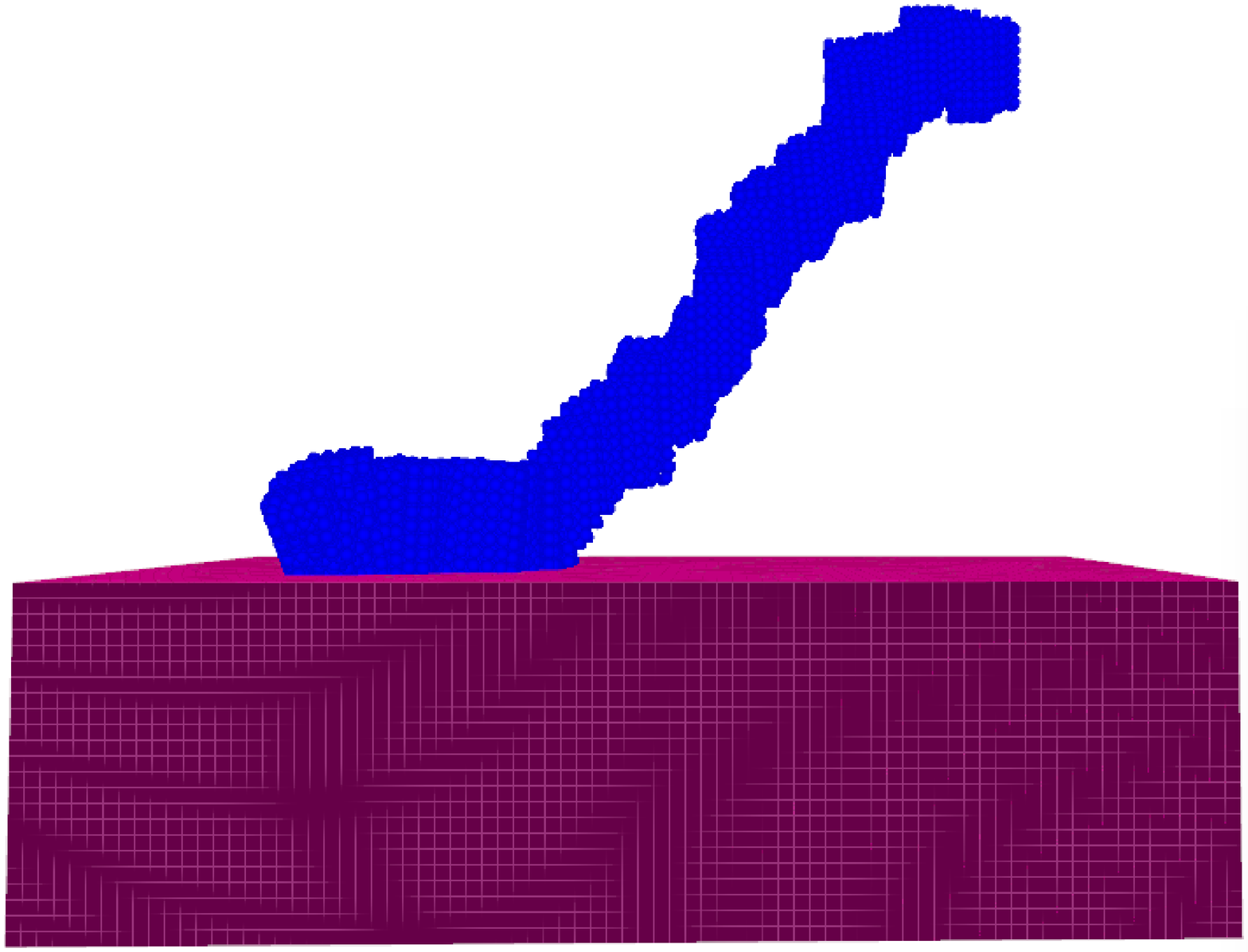}
}
\,
\caption{\small{(a) The $SE(3)PegInHole$ task involves moving from the start (red) to the bottom of the hole. (b) An example policy produced from 296 solutions, the (c) initial action sequence (blue arrows), actions the policy will return if every action is successful, and (d) the swept volume of the peg executing the policy. Note that the peg makes contact with the environment to reduce uncertainty, then slides into the hole.}}
\label{fig:peginholeexamples}
\end{figure*}

\begin{table}[ht!!]
\centering
\begin{tabulary}
{\linewidth}{C|CC|CCCCC}
  \hline
  \hspace{-0.3cm} & AC & PC & \multicolumn{5}{c}{WCR with $D_\mathit{WCR}=$} \\
  \hspace{-0.3cm} & & & \emph{0.125} & \emph{0.25} & \emph{0.5} & \emph{0.75} & \emph{0.99} \\
  \hspace{-0.3cm} $P_\mathit{plan}$ & 1.0 & 1.0 & 1.0 & 0.97 & 0.97 & 0.97 & 0.97 \\
  \hspace{-0.3cm} $P_\mathit{exec}$ & 0.97 [0.17] & 0.89 [0.19] & 0.73 [0.42] & 0.95 [0.18] & 0.84 [0.34] & 0.99 [0.02] & 0.96 [0.18] \\
  \hline
\end{tabulary}
\caption{\small{$SE(3)PegInHole$ particle clustering performance comparison (mean [std.dev.]) of $P_\mathit{exec}$, the probability of reaching the goal with 300 seconds, between policies produced using our planner with different clustering methods. $P_\mathit{plan}$ is the probability that a policy is planned within 5 minutes, averaged over 30 plans, and $P_\mathit{exec}$ is averaged over 40 executions on each successfully-planned policy.}}
\label{tab:clustering}
\end{table}

\begin{table}[ht!!]
\centering
\begin{tabulary}
{\linewidth}{C|CCCC|CCCC}
  \hline
   & \multicolumn{4}{c|}{Simplified} & \multicolumn{4}{c}{Planned policies (WCR, $D_\mathit{WCR}=0.75$)} \\
  \hspace{-0.3cm} & \multicolumn{2}{c}{Simple RRT} & \multicolumn{2}{c|}{Contact RRT} & \multicolumn{2}{c}{24 particles} & \multicolumn{2}{c}{48 particles} \\
  \hspace{-0.3cm} $\gamma$ & $P_\mathit{plan}$ & $P_\mathit{exec}$ & $P_\mathit{plan}$ & $P_\mathit{exec}$ & $P_\mathit{plan}$ & $P_\mathit{exec}$ & $P_\mathit{plan}$ & $P_\mathit{exec}$\\
  \hspace{-0.3cm} 0 & 0 & 0 [0] & 1 & 0.78 [0.38] & 1 & 0.42 [0.48] & 0.97 & 0.59 [0.47] \\
  \hspace{-0.3cm} $\nicefrac{1}{16}$ & 0 & 0 [0] & 1 & 0.78 [0.39] & 1 & 0.60 [0.43] & 1 & 0.625 [0.43] \\
  \hspace{-0.3cm} $\nicefrac{1}{8}$ & 0 & 0 [0] & 1 & 0.79 [0.38] & 1 & 0.99 [0.18] & 0.93 & 0.81 [0.37] \\
  \hspace{-0.3cm} $\nicefrac{1}{4}$ & 0 & 0 [0] & 1 & 0.50 [0.37] & 1 & 0.90 [0.28] & 0.86 & 0.72 [0.41] \\
  \hline
\end{tabulary}
\caption{\small{$SE(3)PegInHole$ policy performance comparison between simplified planners and our planner with 24 and 48 particles and actuation uncertainty $\gamma$.}}
\label{tab:peginholecomparisson}
\end{table}

\subsubsection{Peg-in-hole}
In $SE(3)PegInHole$, a version of the classical peg-in-hole task \cite{LMT} shown in Figure \ref{fig:peginholeexamples}, the free-flying 6-DoF robot ``peg'' must reach the bottom of the hole. This task is difficult for robots with actuation uncertainty, as the hole is only $30\%$ wider than the peg. Even without uncertainty, attempting to avoid contact greatly restricts the motion of the robot entering the hole. Instead, as shown in \cite{LMT}, the best strategy is to use contact with the environment and the compliance of the robot to guide the peg into the hole. We assess the performance of a policy approach in terms of $P_\mathit{exec}$, the probability that executing the policy reaches the goal within a time limit of 300 seconds. For a given value of $\gamma$, linear velocity uncertainty $\gamma_\mathit{v} = \gamma$ (m/s) and angular velocity uncertainty $\gamma_\mathit{\omega} = \nicefrac{1}{4} \gamma$ (rad/s). Linear and angular velocity noise is sample from a zero-mean truncated normal distribution with bounds $[-\gamma_\mathit{v,\omega}, \gamma_\mathit{v,\omega}]$ and standard deviation $\nicefrac{1}{2} \gamma_\mathit{v,\omega}$. While this differs from zero-mean normal distributions conventionally used to model uncertainty, we believe the bounded truncated distribution better reflects the reality of robot actuators, which do not exhibit unbounded velocity error. Goal distance threshold $\epsilon_\mathit{goal}$ was set to $\nicefrac{1}{2}$ the length of the peg.

We first compared the performance of our planner at a fixed $\gamma = \nicefrac{1}{8}$  and $N_\mathit{particles}=24$ using the clustering approaches introduced in Section \ref{sec:particleclustering}, including several thresholds for $D_\mathit{WCR} = {0.125, 0.25, 0.5, 0.75, 0.99}$, with 30 plans per approach (5 minutes planning time) and 40 executions of each planned policy. As seen in Table \ref{tab:clustering}, WCR clustering with $D_\mathit{WCR}=0.75$ clearly outperformed the others in terms of policy success, reaching the goal in $99\%$ of executions. Planning time is overwhelmingly dominated by simulation, accounting for approximately $99.9\%$ of the allotted time. Using WCR and $D_\mathit{WCR}=0.75$, we then compared the performance of our planner against two simplified RRT-based approaches:

\begin{enumerate}
    \item{Simple RRT} --  Does not model uncertainty or allow contact, but like our planner produces multiple solutions in the allotted planning time.
    \item{Contact RRT} -- Incorporates contact and compliance but does not model uncertainty. Equivalent to planning with $\gamma=0$ and one particle.
\end{enumerate}

In addition, we tested our planner with both 24 and 48 particles to show the effects of increasing the number of particles used. As before, we planned 30 policies for each, and executed each planned policy 40 times. Note that the Simple RRT was unable to produce solutions in 5 minutes due to the confined narrow passage. Results are shown in Table \ref{tab:peginholecomparisson}. With low actuator error the Contact RRT performs better, as it does not expend planning time on simulating multiple particles and instead produces more solutions. As error increases, our planner clearly outperforms the alternatives. Note that increasing particles does not improve performance, indicating that 24 particles is sufficient without requiring unnecessary simulation. Low $P_\mathit{exec}$ overall, in particular when planning with low $\gamma$, is due to the mismatch between the planning simulator and the dynamics of Gazebo (i.e. motions that are possible in the planner, but not in Gazebo) which disproportionally affects motions near the entrance to the hole. In particular, the planner at low values of $\gamma$ overestimates how successfully motions at the edge of the hole can be performed and thus results in a lower-than-expected $P_\mathit{exec}$.

In terms of the number of particles stored, the worst case was $N_\mathit{particles}=48, \gamma=0$, with an average of 148894 (std.dev. 25015) particles stored. The worst case for simulated particles was $N_\mathit{particles}=24, \gamma=\nicefrac{1}{4}$, with an average 268634 (std.dev. 16856) particles simulated. In practice, the storage and computational expense is limited; the worst-case for particles stored requires a mere 15 megabytes, while for a planning time of 300 seconds and using eight threads, the planner evaluated more than 100 particles per second per thread.

\begin{figure*}[ht!!!]
\centering
\subfloat[]{\label{fig:baxters1}
\includegraphics[width=0.2544\textwidth]{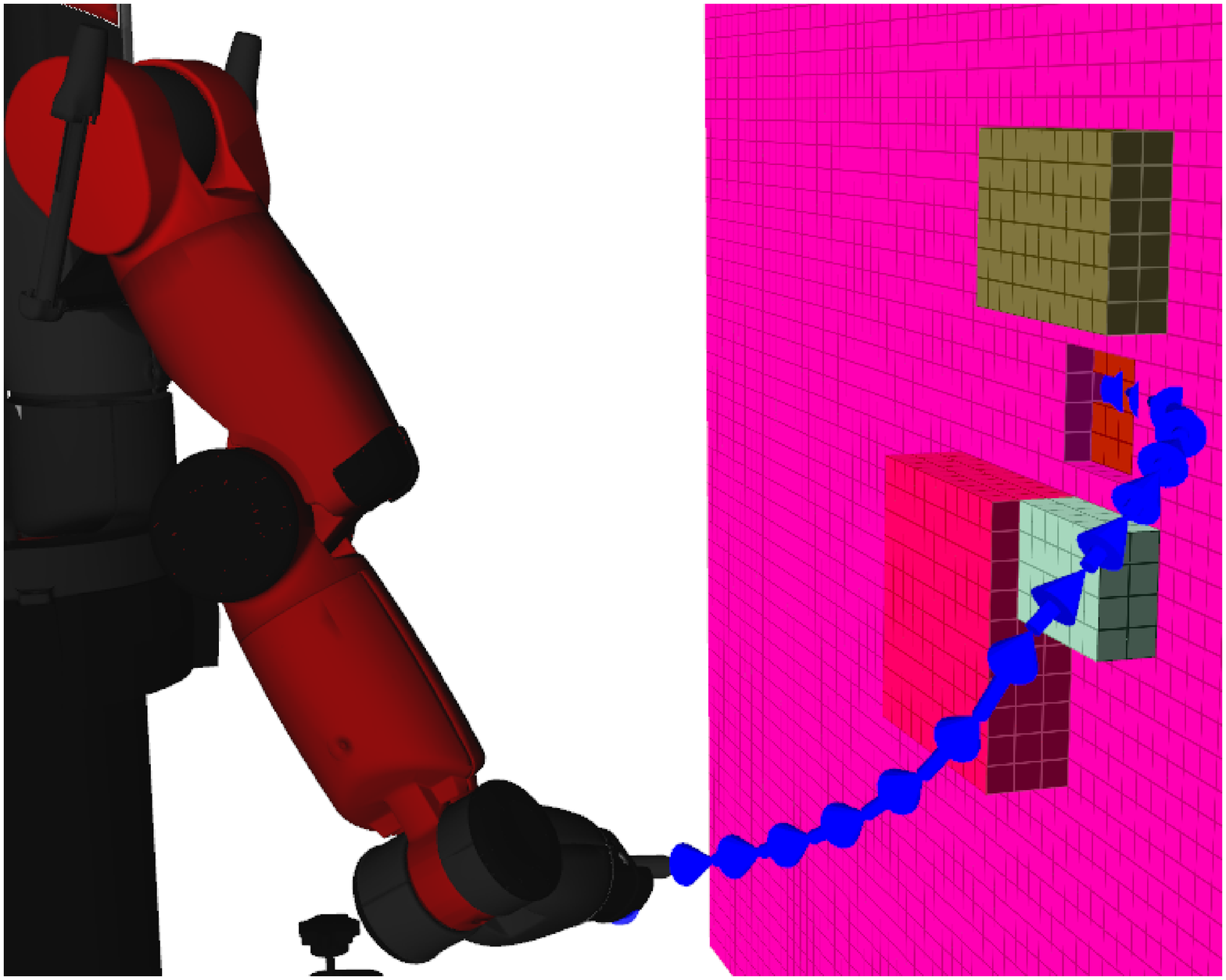}
}
\subfloat[]{\label{fig:baxters2}
\includegraphics[width=0.2355\textwidth]{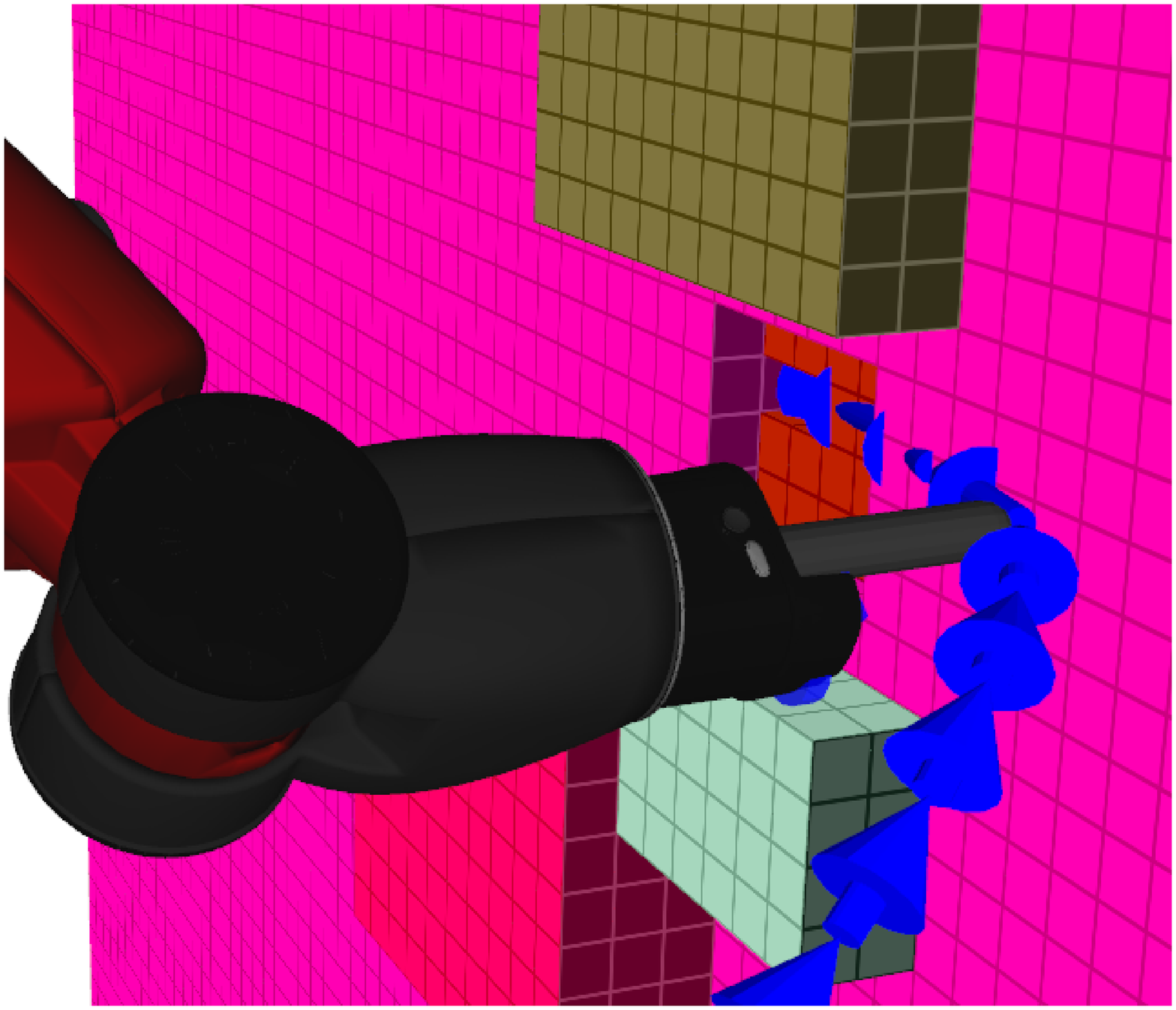}
}
\subfloat[]{\label{fig:baxters3}
\includegraphics[width=0.22425\textwidth]{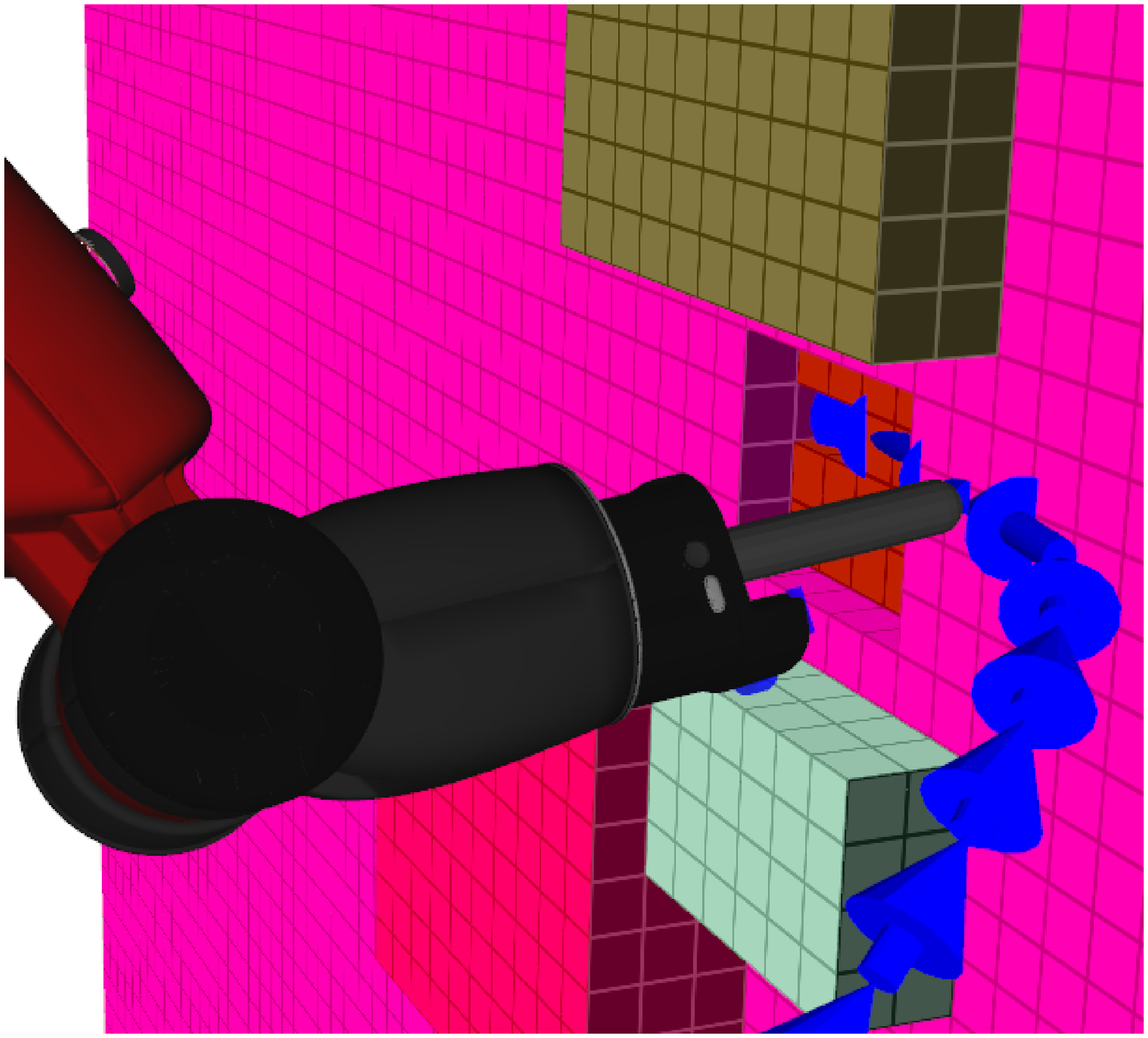}
}
\subfloat[]{\label{fig:baxters4}
\includegraphics[width=0.21984\textwidth]{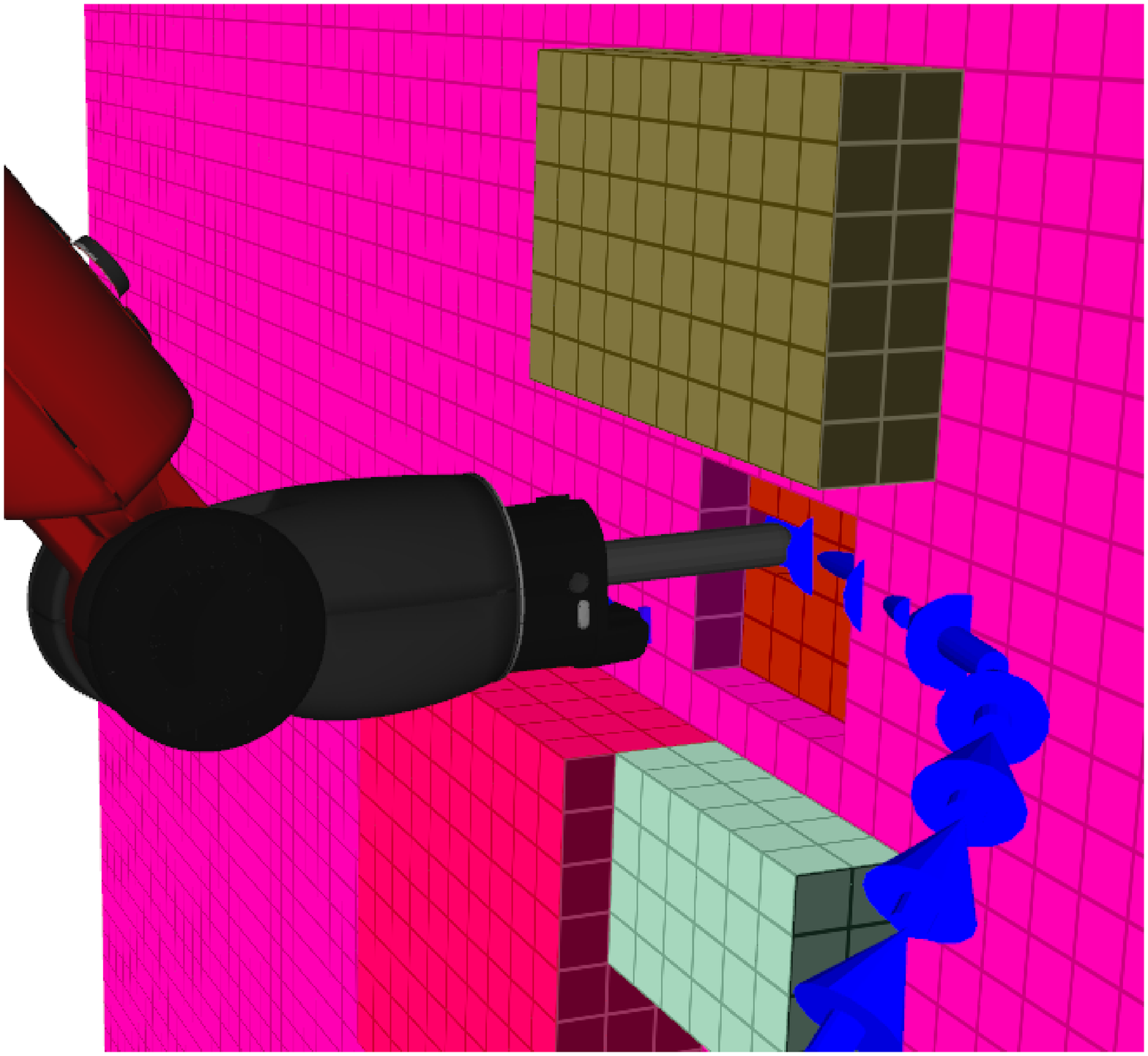}
}
\,
\caption{\small{An execution of the Baxter task, from start (a) to goal (d), and environment with confined space around the goal. The planned policy is shown in blue. Note the use of contact with the environment to reduce uncertainty and reach the target passage.}}
\label{fig:baxterexamples}
\end{figure*}

\subsection{Baxter simulation}
\label{sec:BaxterSimResults}
In addition to $SE(3)$ and $SE(2)$ tests, we apply our planner and policy execution to a simulated Baxter robot shown in Figure \ref{fig:baxterexamples}, with the robot reaching into a confined space. We compare the performance of our planner with $N_\mathit{particles}=24$ and WCR clustering method with $D_\mathit{WCR}=0.1$ with the simplified Contact RRT in terms of success probability $P_\mathit{exec}$ for uncertainty $\gamma=0.1$. To simulate Baxter's actuation uncertainty $\gamma$ defines a truncated normal distribution with $\sigma=\nicefrac{1}{2}\gamma\dot{q_\mathit{i}}$ and bounds $[-\gamma\dot{q_\mathit{i}},\gamma\dot{q_\mathit{i}}]$ for each joint $i$ with velocity $\dot{q_\mathit{i}}$. Goal distance threshold $\epsilon_\mathit{goal} = 0.15$ radians. We generated 10 policies using each approach with a planning time of 10 minutes to ensure both approaches would produce multiple solutions, then executed each 8 times for up to 5 minutes. As expected, Contact RRT finds solutions faster; on average 8.42s (std.dev. 2.61) versus 65.4s (std.dev. 32.9) and policies contain more solutions; on average 17.2 (std.dev. 16.5) versus 6.33 (std.dev. 3.97) since each solution requires less simulation time. However, our planner incorporating uncertainty outperforms the Contact RRT baseline with $P_\mathit{exec}=0.79$ (std.dev. 0.30) versus $P_\mathit{exec}=0.70$ (std.dev. 0.30). This suggests that, while planning with uncertainty does help in this environment, our approach to policy execution and resilience also works well when uncertainty is not accounted for in the planner, but we have a diverse policy.



\subsection{Policy adaptation}

\begin{figure*}[ht!!]
\centering
\subfloat[]{\label{fig:planarexamplepolicy}
\includegraphics[width=0.23\textwidth]{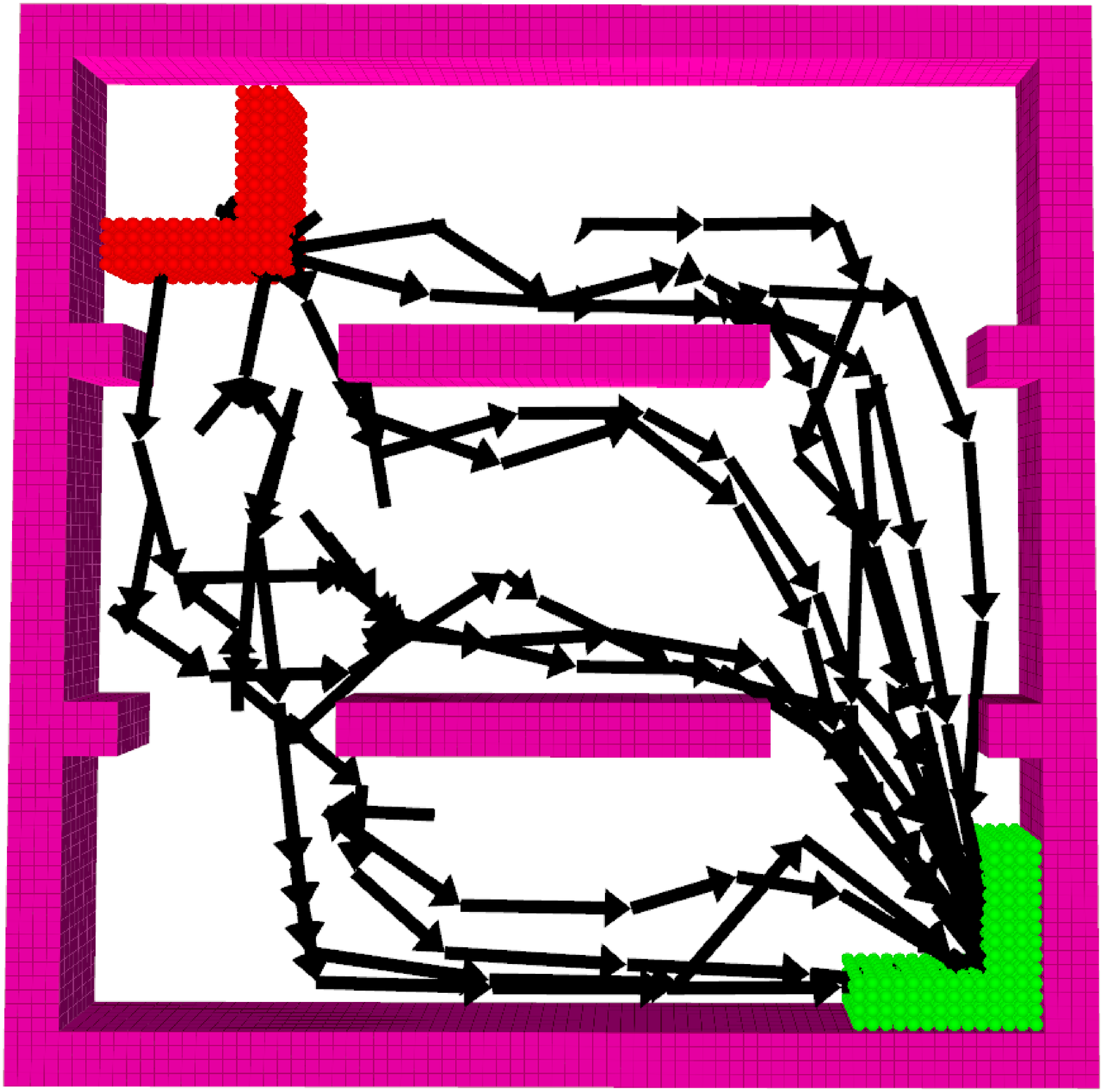}
}
\subfloat[]{\label{fig:initialpolicy}
\includegraphics[width=0.23\textwidth]{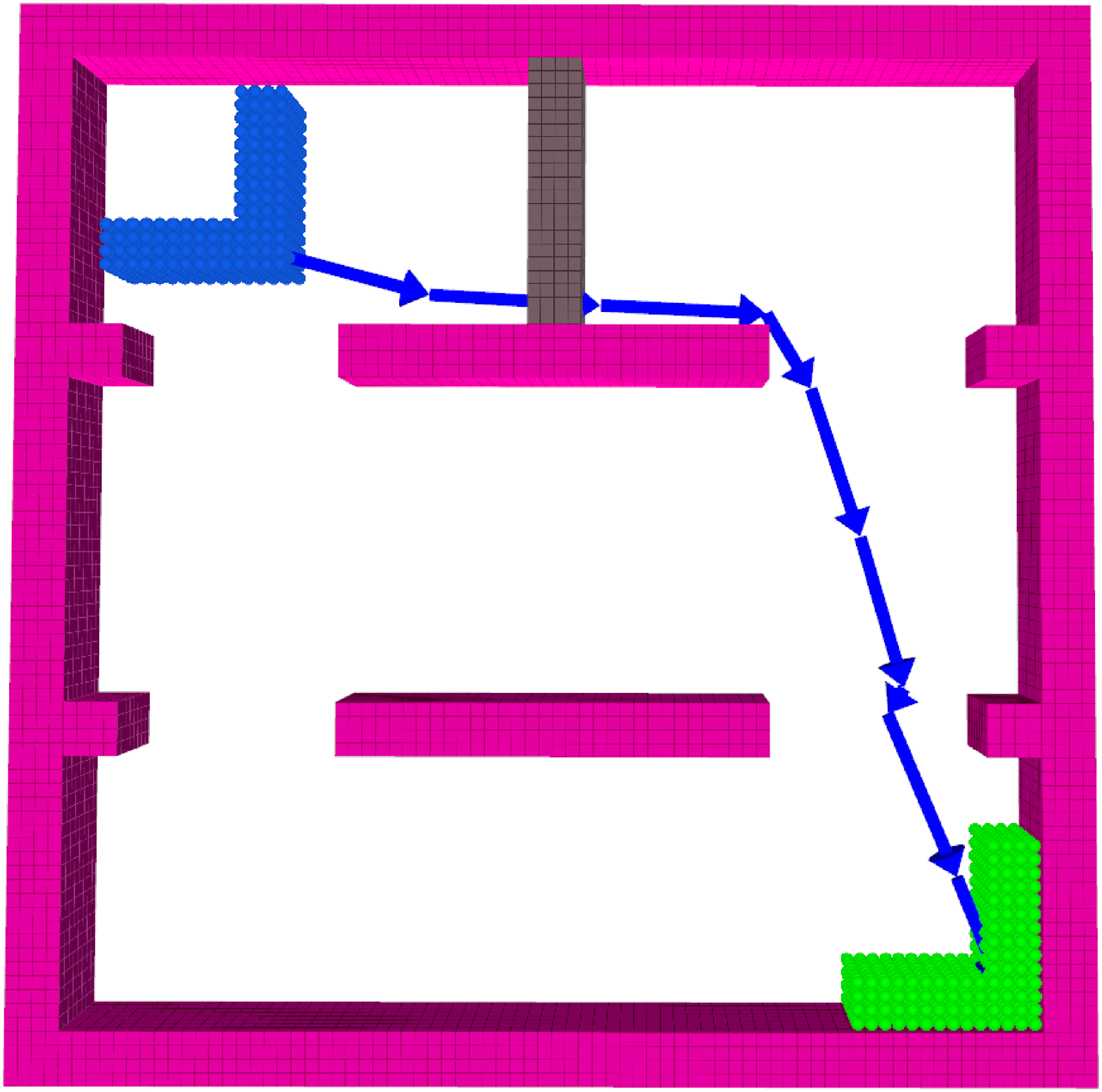}
}
\subfloat[]{\label{fig:policycollided}
\includegraphics[width=0.23\textwidth]{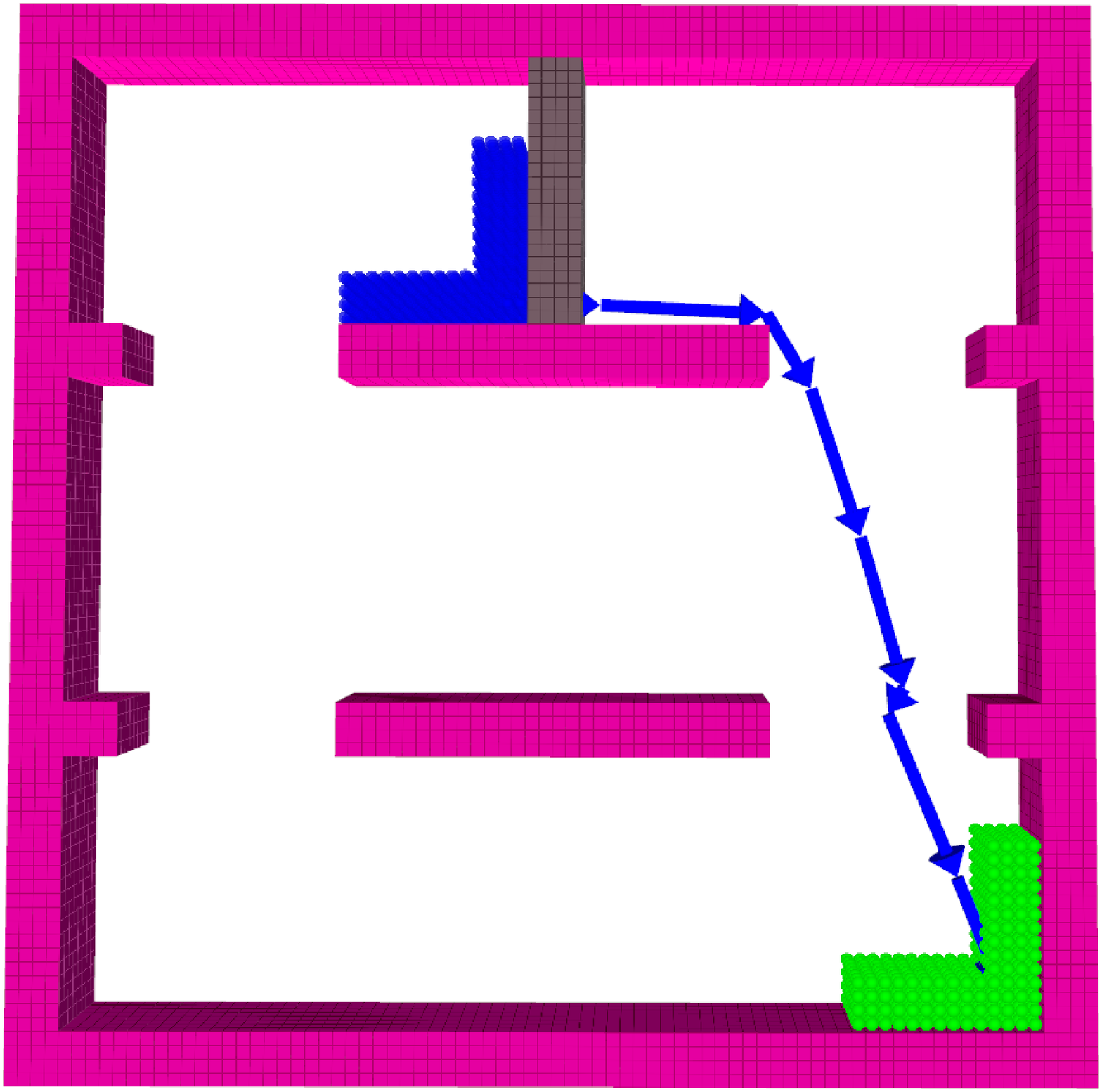}
}
\subfloat[]{\label{fig:updatedpolicy}
\includegraphics[width=0.23\textwidth]{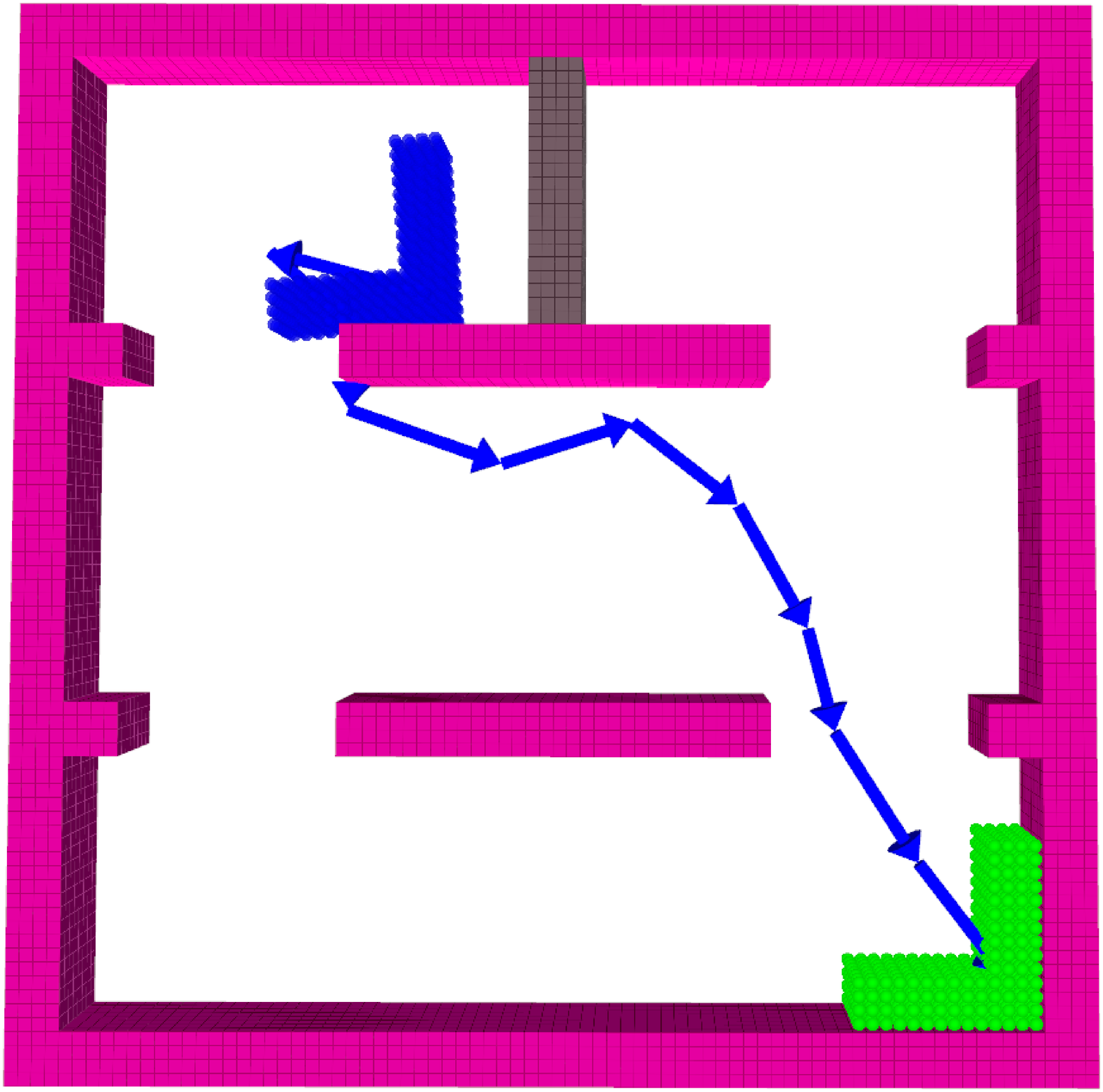}
}
\,
\caption{\small{(a) Our planar test environment, in which the robot must move from upper left (red) to lower right (green), with an example policy produced by our planner, with solutions through each of the horizontal passages. (b) The initial action sequence (blue arrows), showing actions the policy will return if every action is successful. (c) Following the policy, the robot becomes stuck on the new obstacle (gray). (d) Once the policy detects the failed action, it adapts to avoid the obstacle.}}
\label{fig:planartestenv}
\vspace{-0.5cm}
\end{figure*}

We performed tests in a planar $SE(2)$ (3-DoF) environment to show that our policy adaptation recovers from unexpected behavior during execution. As shown in Figure \ref{fig:planartestenv}, the L-shaped robot attempts to move from the start (upper left) to the goal (lower right). Due to the obstacles present, there are three distinct horizontal passages. Using the same controllers and uncertainty models as the $SE(3)$ tests with uncertainty $\gamma = 0.125$ and WCR clustering method with $D_\mathit{WCR}=0.75$, 24 particles, and a planning time of 2 minutes, we generated 30 policies using our planner. Goal distance threshold $\epsilon_\mathit{goal}$ was set to $\nicefrac{1}{8}$ the length of the robot.

We evaluated the performance of the planned policies in the unmodified environment and an environment in which we blocked the horizontal passage used by the initial path of each policy. Each was executed 8 times for a total of 240 policy executions, for a maximum of 600 seconds. In the unmodified environment, $97\%$ of policies were executed successfully, with an average of 15.4 actions (std.dev. 9.62). In the modified environment with policy execution importance $A_\mathit{importance} = 500$ (this high value results in rapid policy adaptation), $73\%$ of policies were executed successfully, with an average of 26.7 actions (std.dev. 12.3). 
This result shows that adapting the policy using our methods allows us to circumvent the new obstacle, however, doing so may result in following a path that is less likely to succeed. 

\vspace{-0.1cm}

\section{Conclusion}
\label{sec:conclusion}

We have developed a method for planning motion for robots with actuation uncertainty that incorporates environment contact and compliance of the robot to reliably perform manipulation tasks. First, we generate partial policies using an RRT-based motion planner that uses particle-based models of uncertainty and kinematic simulation of contact and compliance. Second, we adapt planned policies online during execution to account for unexpected behavior that arises from model or environment inaccuracy. We have tested our planner and policy execution in simulated $SE(2)$ and $SE(3)$ environments and on the simulated Baxter robot and show that our methods generate policies that perform manipulation tasks involving significant contact and compare against two simpler methods. Additionally, we show that our policy adaptation is resilient to significant changes during execution; e.g. adding a new obstacle to the environment.

\subsubsection{Acknowledgements}
This work was supported in part by NSF grants IIS-1656101 and IIS-1551219.

\bibliographystyle{splncs}

\bibliography{references}

\begin{thebibliography}{10}

\bibitem{LMT}
Lozano-Pérez, T., Mason, M.T., Taylor, R.H.:
\newblock Automatic synthesis of fine-motion strategies for robots.
\newblock IJRR \textbf{3}(1) (1984)  3--24

\bibitem{conformant}
Goldman, R.P., Boddy, M.S.:
\newblock Expressive planning and explicit knowledge.
\newblock In: Artificial Intelligence Planning Systems. (May 1996)

\bibitem{particleRRT}
Melchior, N.A., Simmons, R.:
\newblock Particle rrt for path planning with uncertainty.
\newblock In: ICRA. (April 2007)

\bibitem{finemotioncomputability}
Canny, J.:
\newblock On computability of fine motion plans.
\newblock In: ICRA. (May 1989)

\bibitem{backprojections}
Erdmann, M.:
\newblock Using backprojections for fine motion planning with uncertainty.
\newblock The International Journal of Robotics Research \textbf{5}(1) (1986)
  19--45

\bibitem{SARSOP}
Kurniawati, H., Hsu, D., Lee, W.S.:
\newblock Sarsop: Efficient point-based pomdp planning by approximating
  optimally reachable belief spaces.
\newblock In: RSS. (2008)

\bibitem{MCVI}
Bai, H., Hsu, D., Kochenderfer, M., Lee, W.S.:
\newblock Unmanned aircraft collision avoidance using continuous-state pomdps.
\newblock In: RSS. (June 2011)

\bibitem{contactpomdp}
Koval, M., Pollard, N., Srinivasa, S.:
\newblock Pre- and post-contact policy decomposition for planar contact
  manipulation under uncertainty.
\newblock In: RSS. (July 2014)

\bibitem{policysearch}
Levine, S., Wagener, N., Abbeel, P.:
\newblock Learning contact-rich manipulation skills with guided policy search.
\newblock In: ICRA. (May 2015)

\bibitem{beliefroadmap}
Roy, N., Prentice, S.:
\newblock The belief roadmap: Efficient planning in belief space by factoring
  the covariance.
\newblock IJRR \textbf{28}(11-12) (2009)  1448--1465

\bibitem{RRBT}
Bry, A., Roy, N.:
\newblock Rapidly-exploring random belief trees for motion planning under
  uncertainty.
\newblock In: ICRA. (May 2011)

\bibitem{FIRM}
Agha-mohammadi, A.a., Chakravorty, S., Amato, N.M.:
\newblock Firm: Sampling-based feedback motion planning under motion
  uncertainty and imperfect measurements.
\newblock The International Journal of Robotics Research (2013)

\bibitem{stochasticmotionroadmap}
Alterovitz, R., Siméon, T., Goldberg, K.:
\newblock The stochastic motion roadmap: A sampling framework for planning with
  markov motion uncertainty.
\newblock In: RSS. (June 2007)

\bibitem{beliefdistance}
Littlefield, Z., Klimenko, D., Kurniawati, H., Bekris, K.E.:
\newblock The importance of a suitable distance function in belief-space
  planning.
\newblock In: ISRR. (September 2015)

\bibitem{LQGMP}
Berg, J.V.D., Abbeel, P., Goldberg, K.:
\newblock Lqg-mp: Optimized path planning for robots with motion uncertainty
  and imperfect state information.
\newblock In: RSS. (June 2010)

\bibitem{incrementastochastic}
Huynh, V.A., Karaman, S., Frazzoli, E.:
\newblock An incremental sampling-based algorithm for stochastic optimal
  contro.
\newblock In: ICRA. (May 2012)

\bibitem{copt}
Davis, B., Karamouzas, I., Guy, S.J.:
\newblock C-opt: Coverage-aware trajectory optimization under uncertainty.
\newblock IEEE Robotics and Automation Letters \textbf{1}(2) (July 2016)
  1020--1027

\bibitem{sigmahulls}
Lee, A., Duan, Y., Patil, S., Schulman, J., McCarthy, Z., van~den Berg, J.,
  Goldberg, K., Abbeel, P.:
\newblock Sigma hulls for gaussian belief space planning for imprecise
  articulated robots amid obstacles.
\newblock In: IROS. (Nov 2013)

\bibitem{pushingRRT}
Nieuwenhuisen, D., van~der Stappen, A.F., Overmars, M.H.:
\newblock Pushing using compliance.
\newblock In: ICRA. (May 2006)

\bibitem{kinodynamicRRT}
LaValle, S.M., Kuffner, J.J.:
\newblock Randomized kinodynamic planning.
\newblock IJRR \textbf{20}(5) (2001)  378--400

\bibitem{numericaltaxonomy}
Sneath, P.H.A., Sokal, R.R.:
\newblock Numerical taxonomy: the principles and practice of numerical
  classification.
\newblock Freeman (1973)

\bibitem{weakconvex}
Asafi, S., Goren, A., Cohen-Or, D.:
\newblock Weak convex decomposition by lines-of-sight.
\newblock Computer Graphics Forum \textbf{32}(5) (2013)  23--31

\end{thebibliography}

\newpage

\begin{subappendices}
\renewcommand{\thesection}{\Alph{section}}

\section{Fast kinematic simulation}
\label{app:kinematicsimulation}

\begin{figure*}[ht!]
    \centering
    \subfloat[a]{\label{fig:pre_resolve}\includegraphics[width=0.2\textwidth]{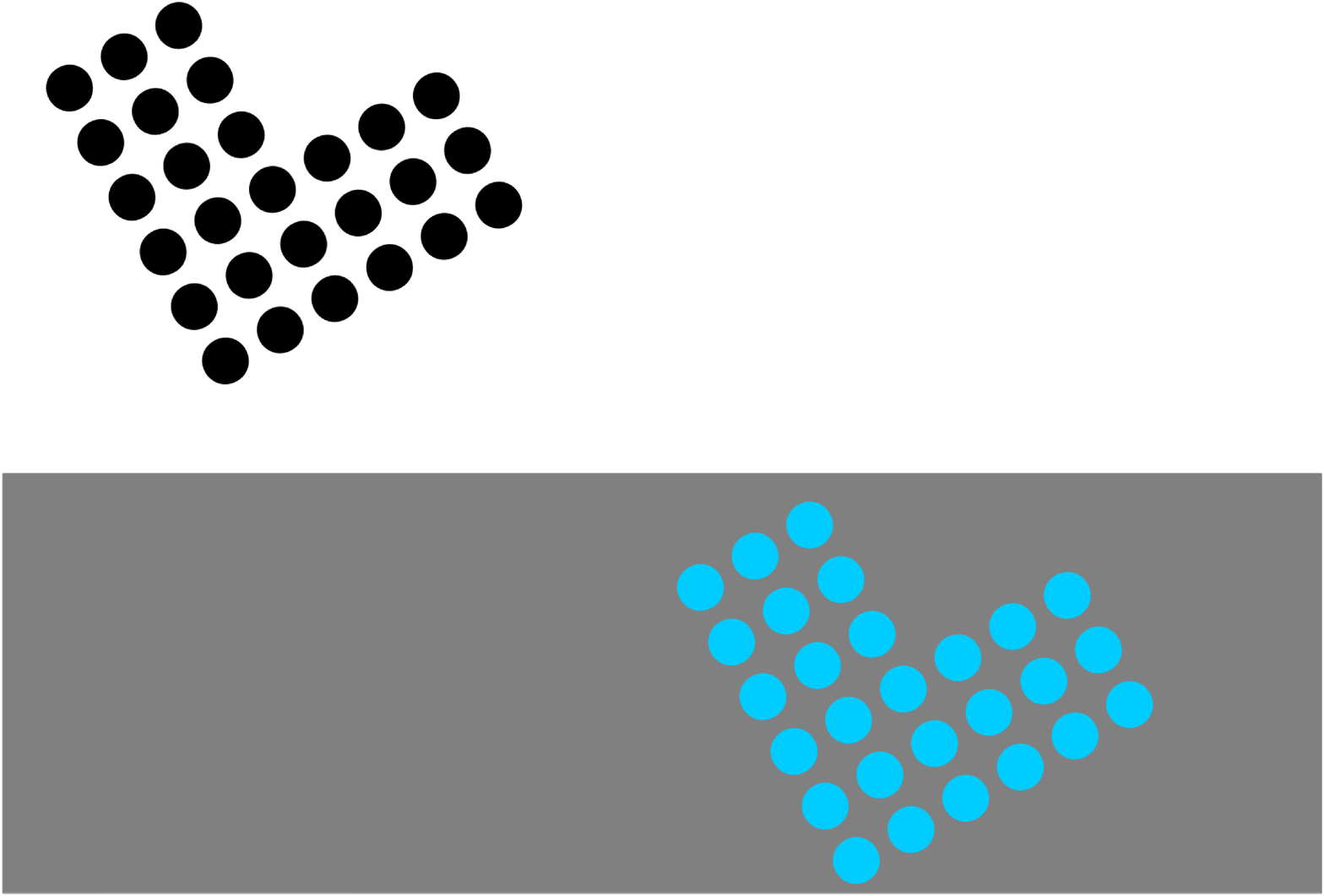}
    }
    \subfloat[b]{\label{fig:resolve_triggered}\includegraphics[width=0.2\textwidth]{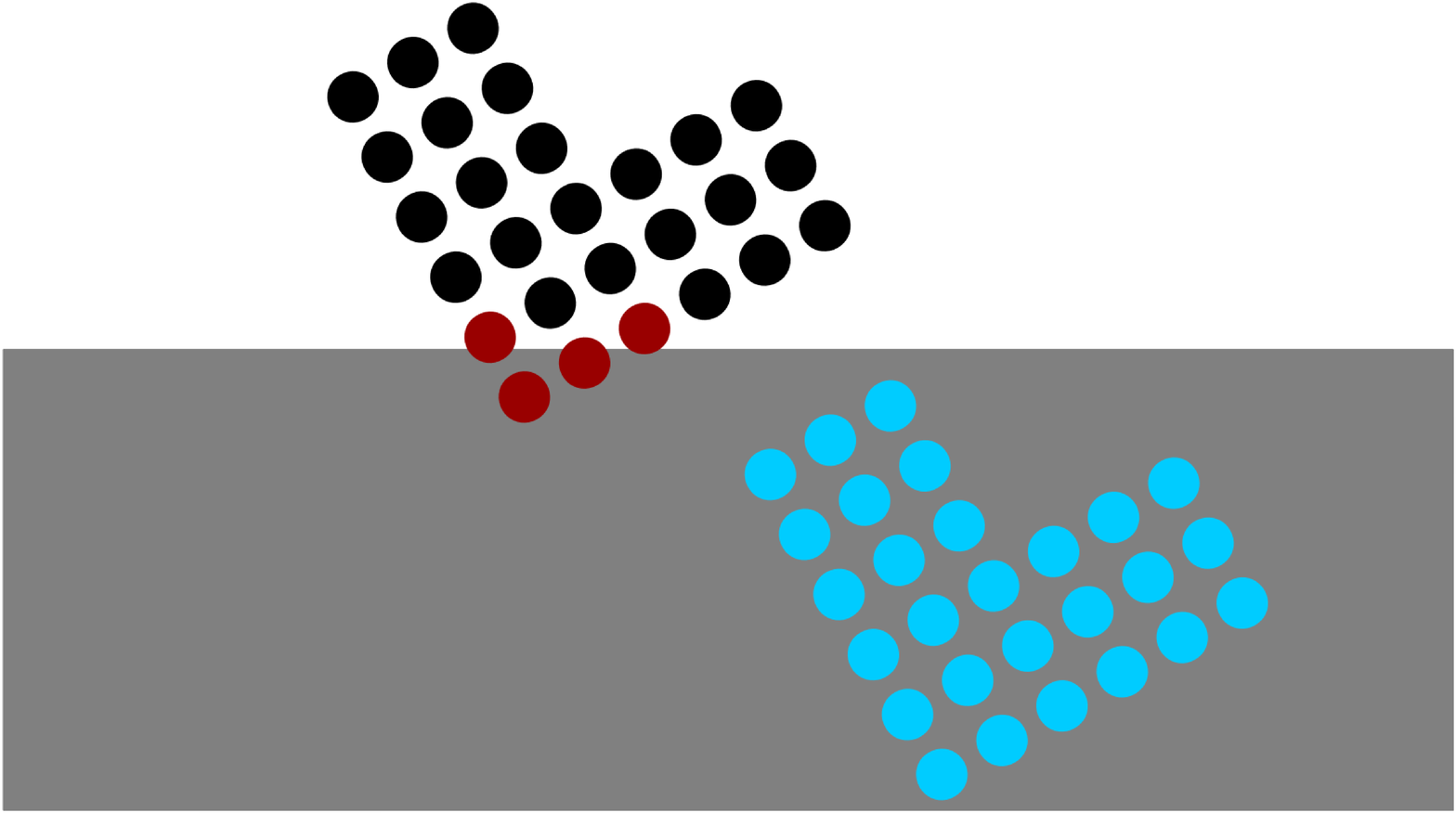}
    }
    \subfloat[c]{\label{fig:resolve_deltas}\includegraphics[width=0.2\textwidth]{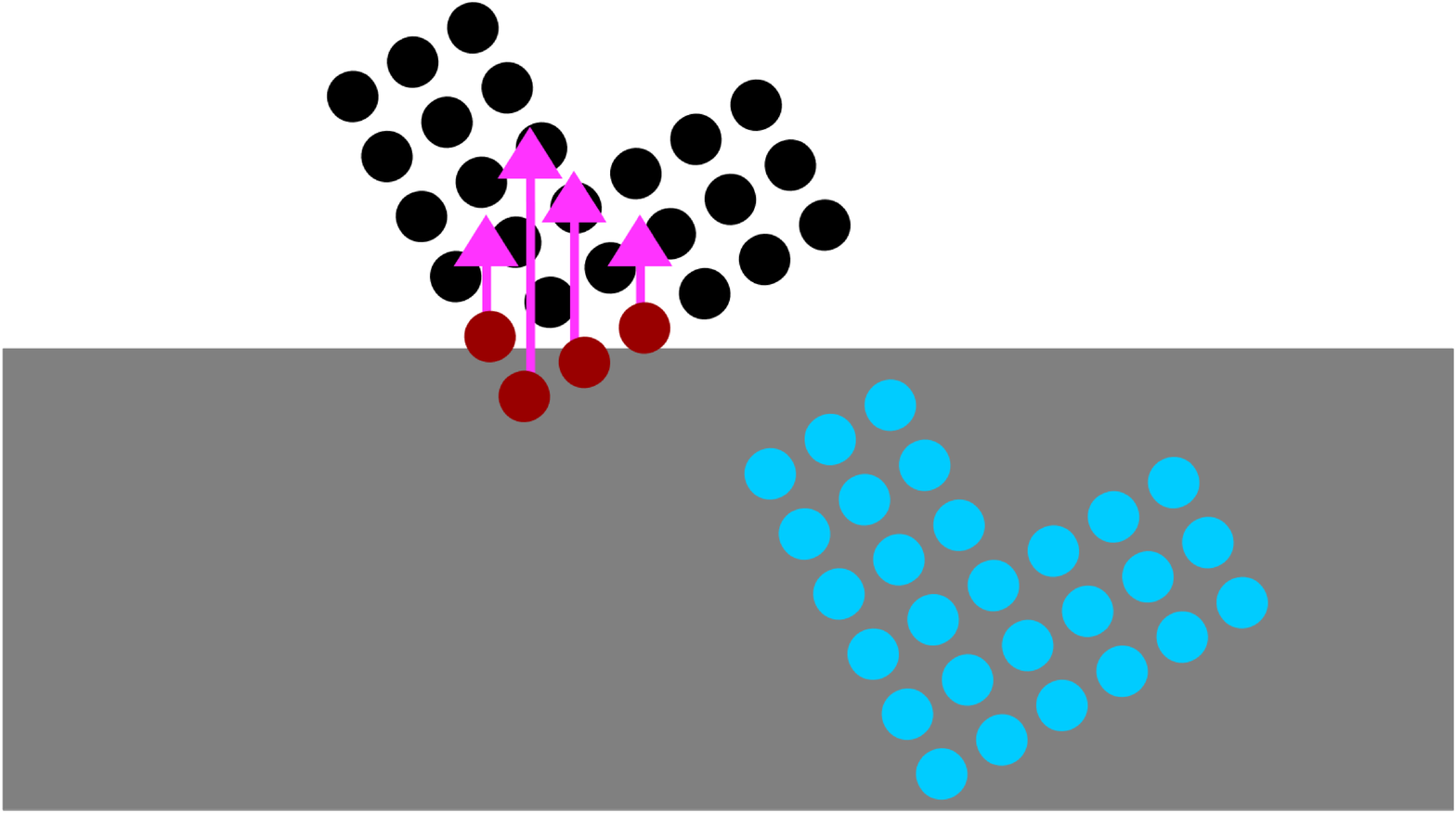}
    }
    \subfloat[d]{\label{fig:post_resolve}\includegraphics[width=0.2\textwidth]{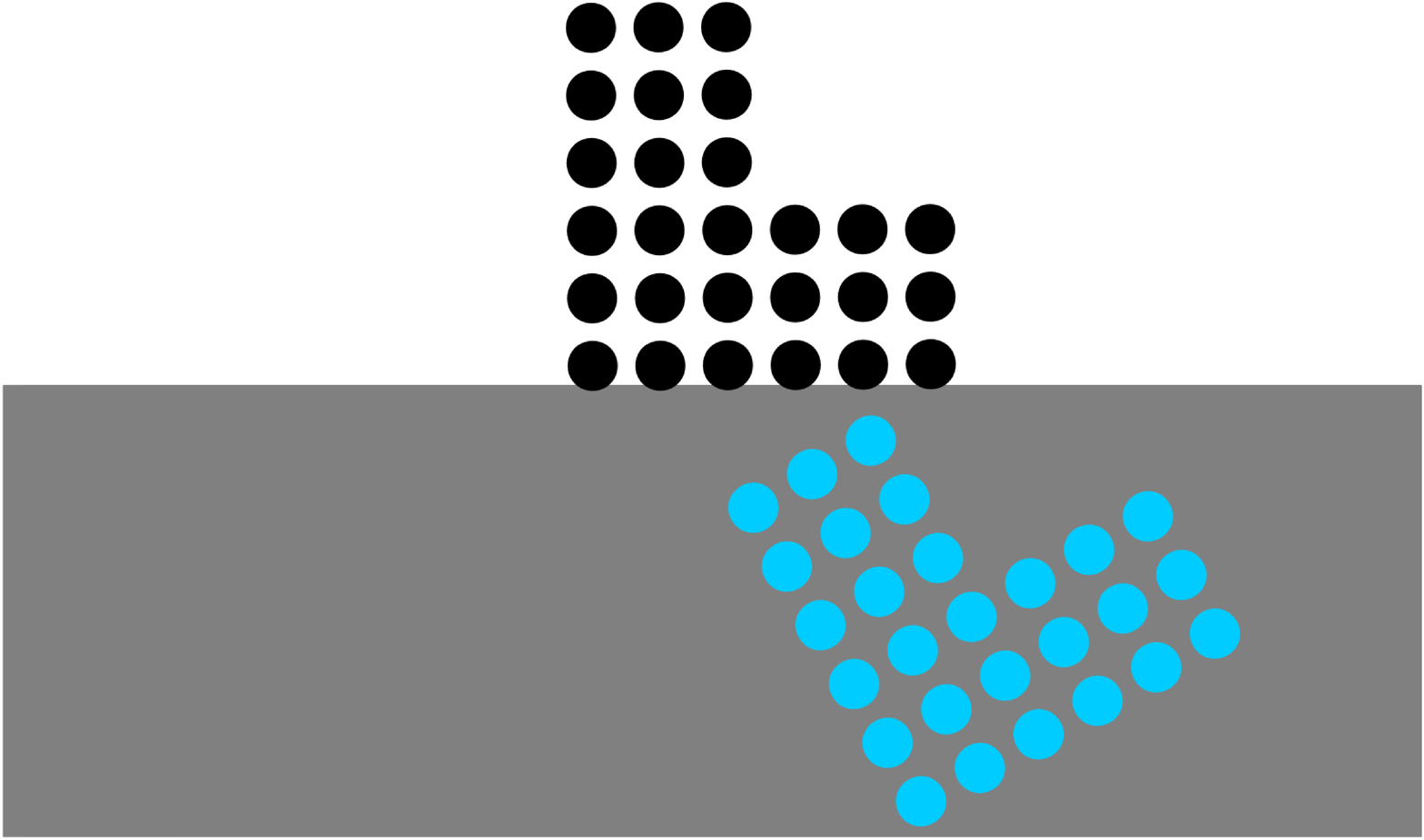}
    }
    \,
    \caption{\small{The collision resolution process used by our lightweight simulator. From left to right, (a) a robot represented by points (black) moving towards a target (light blue) and an obstacle (gray) (b) collides with the object, triggering the collision resolution. (c) point corrections $\Delta p_n$ for each colliding point of the robot are computed from the surface normals of the object and applied (d) so the robot complies to produce an in-contact state.}}
    \label{fig:collisionresolution}
\end{figure*}

The global and local planners rely on the presence of a computationally-efficient simulator for the behavior of our controlled compliant robot. While progress has been made in the performance of dynamic simulators, we require a simulator capable of evaluating hundreds, if not thousands, of robot motions per second to grow the planner's tree in reasonable time. To improve computational performance, we limit ourselves to kinematic simulation, though our planning framework is agnostic to the simulator being used. Kinematic simulation is only an approximation of true robot motion; however, we mitigate the discrepancy between simulated and real dynamic behavior using policy adaptation to update the planned policy with the results of real-world executions. For this work, the robot is controlled via PD feedback controller with gains $K_\mathit{p},K_\mathit{d}$, for error $e_\mathit{q} = q_\mathit{desired} - q$ the resulting control input is $\Delta q = K_\mathit{p} e + K_\mathit{d} \dot{e_\mathit{q}}$; however, different controllers can be used with the simulator. For a fixed time limit $t_\mathit{simulate}$, we forward simulate the motion of the robot from the current configuration $q_\mathit{t}$ to the next configuration $q_\mathit{t+1}$ using the equation below.

\begin{align}
\label{eq:simulation}
e_\mathit{q} = q_\mathit{target} - q_\mathit{t} \\
\Delta q = K_p e_\mathit{q} + K_d \dot{e_\mathit{q}} \\
q_\mathit{t+1} = q + \Delta q + \bold{F}(\Delta q) \\
q_\mathit{t+1}' = \Call{ResolveCollisions}{q_\mathit{t+1}}
\end{align}

Collision resolution and robot compliance are modelled by $\Call{ResolveCollisions}{}$, which iteratively corrects colliding configurations $q_\mathit{t+1}$ until an in-contact configuration $q_\mathit{t+1}'$ is reached. For performance purposes, the environment $E$ is modelled using a voxel grid that stores the surface normals for all obstacles in $E$, and the robot $R$ is modelled using a set of points for each link. Collision checking of a configuration $q$ is performed by transforming every point of the robot into the environment and checking if any of the corresponding voxels belong to an obstacle. If any voxels belong to an obstacle, the collision is resolved by iteratively applying corrections $\Delta q$ as shown in Figure \ref{fig:collisionresolution}. Each correction $\Delta q$ is the product of the individual point corrections $\Delta p_n$ for each colliding point, where $\Delta p_n$ is the product of the surface normal of the collided obstacle and penetration distance of point $p_n$, and the Jacobian $\bold{J}$ pseudoinverse of the robot for configuration $q$ and point $p_n$ as shown below:

\begin{align}
\label{eq:resolution}
\Delta q &= \textbf{J}(q,p_1,p_2,...)^+ [\Delta p_1^T, \Delta p_2^T,...]^T
\end{align}

Intuitively, this computes the change in configuration necessary to move points $p_1, p_2,...$ out of collision, where the correct direction to move of out of collision is approximated by the surface normal of the collided object. Note that this approximation is only valid if the maximum penetration of an obstacle is small; thus we use small timesteps in both $\Call{ForwardSimulate}{}$ and $\Call{ResolveCollisions}{}$ to ensure that the workspace motion of the robot is small. While our kinematic simulation does not consider surface friction which could hamper sliding motions, we address this using a simple controller discussed in Section \ref{app:dynamicsimulation} and our simulation results show that this limitation does not overly impair the performance of our planner, though unexpected jamming could still occur.

\section{Execution controllers}
\label{app:dynamicsimulation}

\begin{figure*}[ht!]
    \centering
    \subfloat[a]{\label{fig:robotabouttocollide}\includegraphics[width=0.23\textwidth]{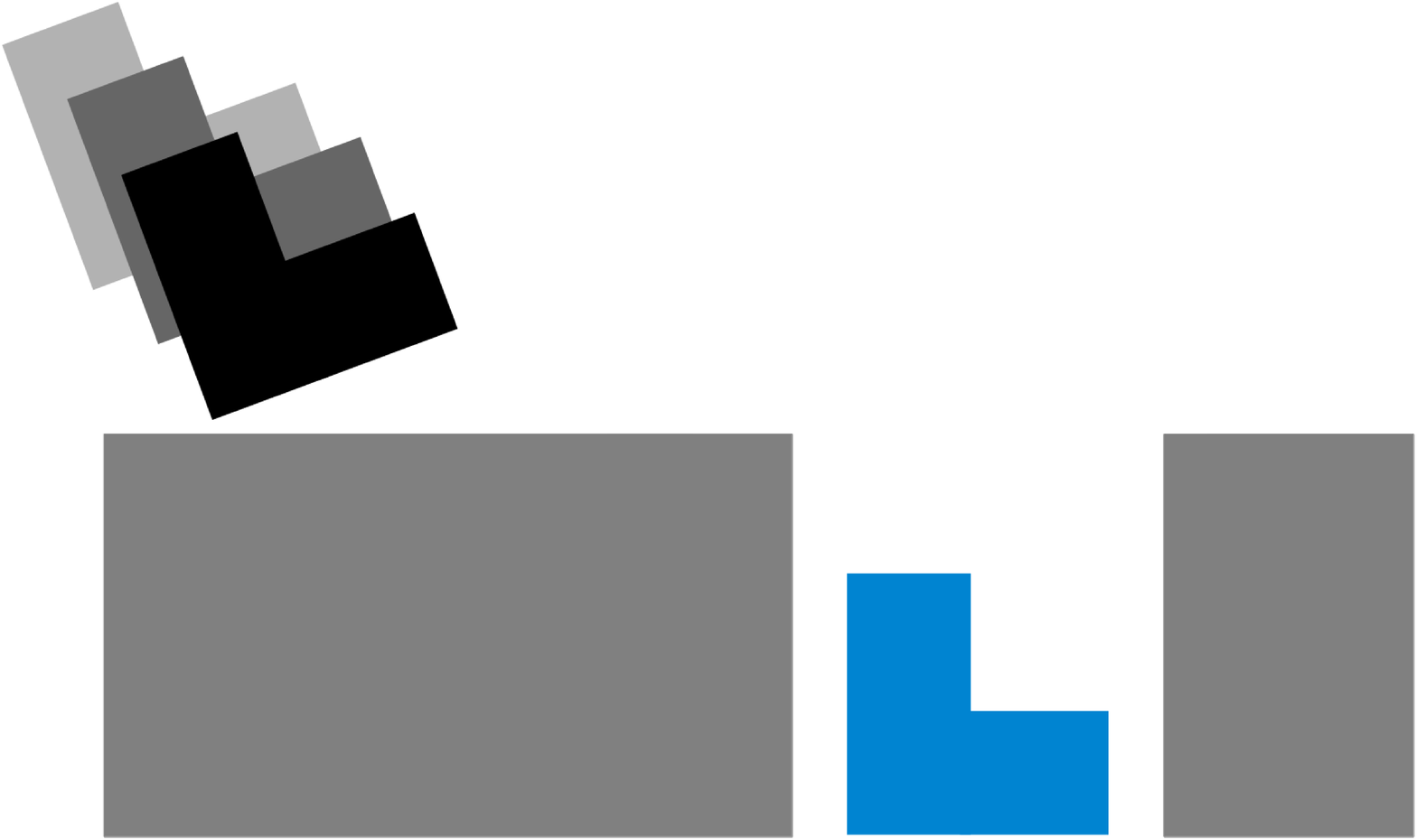}
    }
    \subfloat[b]{\label{fig:robotincontact}\includegraphics[width=0.23\textwidth]{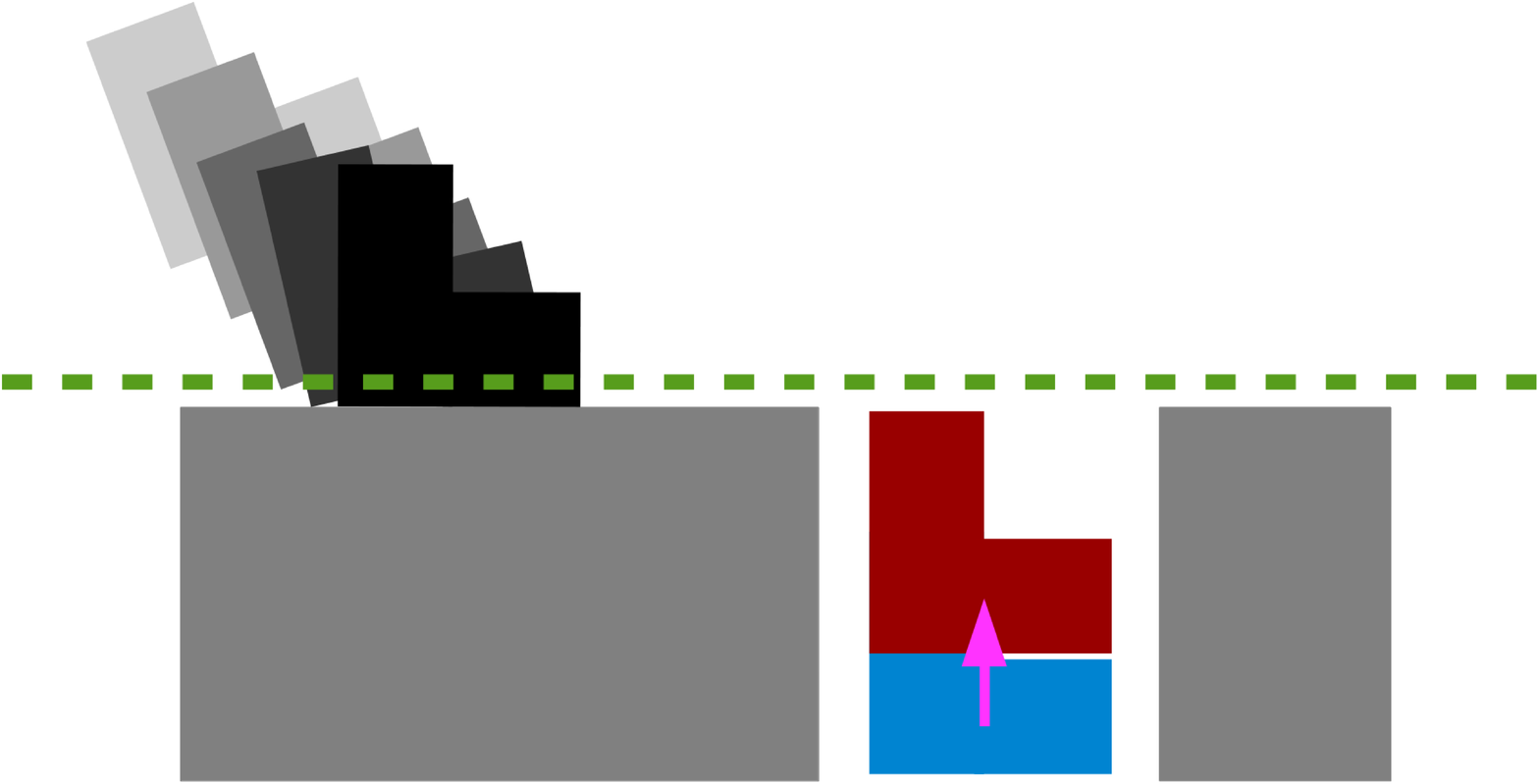}
    }
    \subfloat[c]{\label{fig:robotmovingagain}\includegraphics[width=0.23\textwidth]{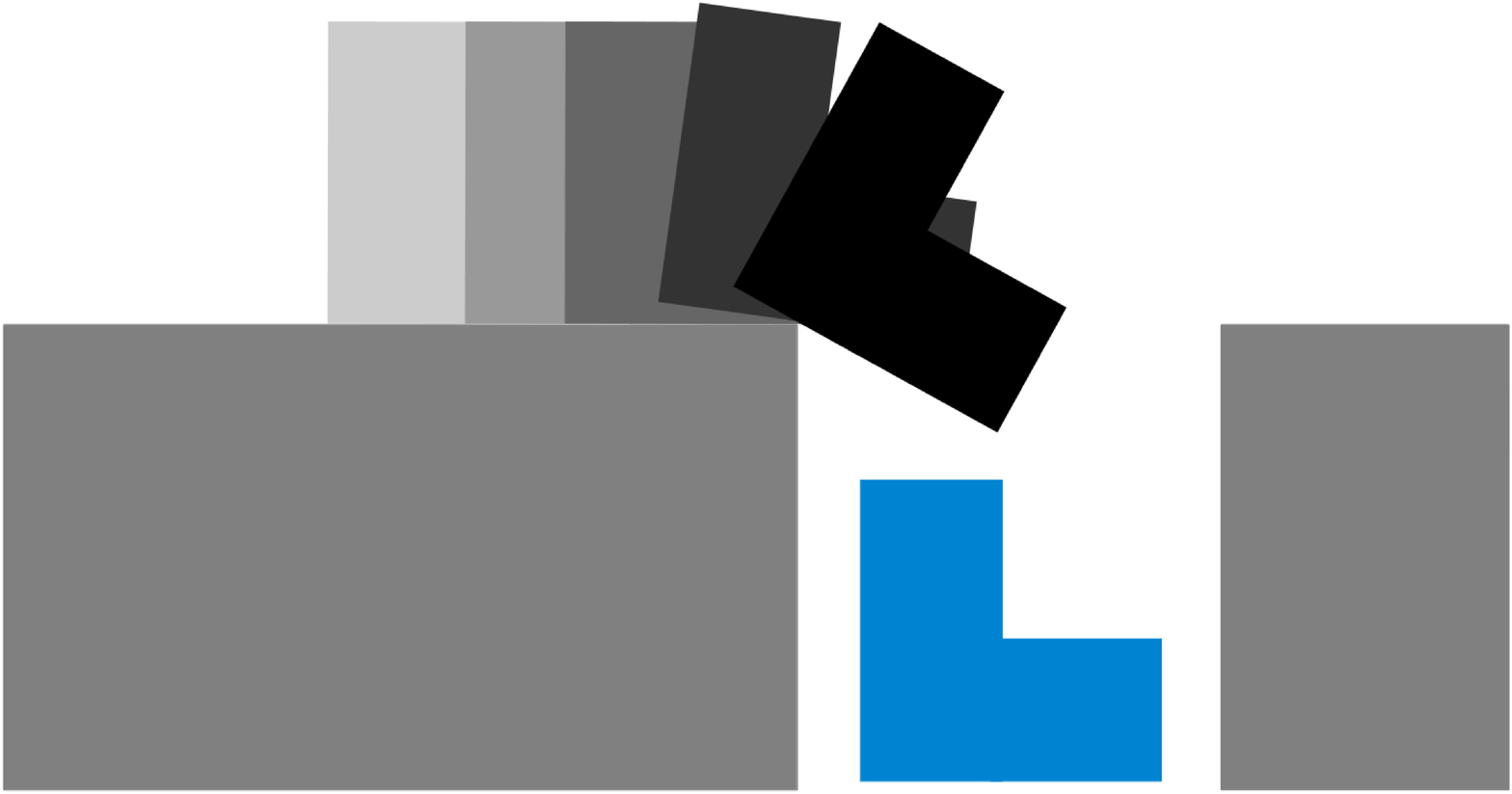}
    }
    \subfloat[d]{\label{fig:completelystuckrobot}\includegraphics[width=0.23\textwidth]{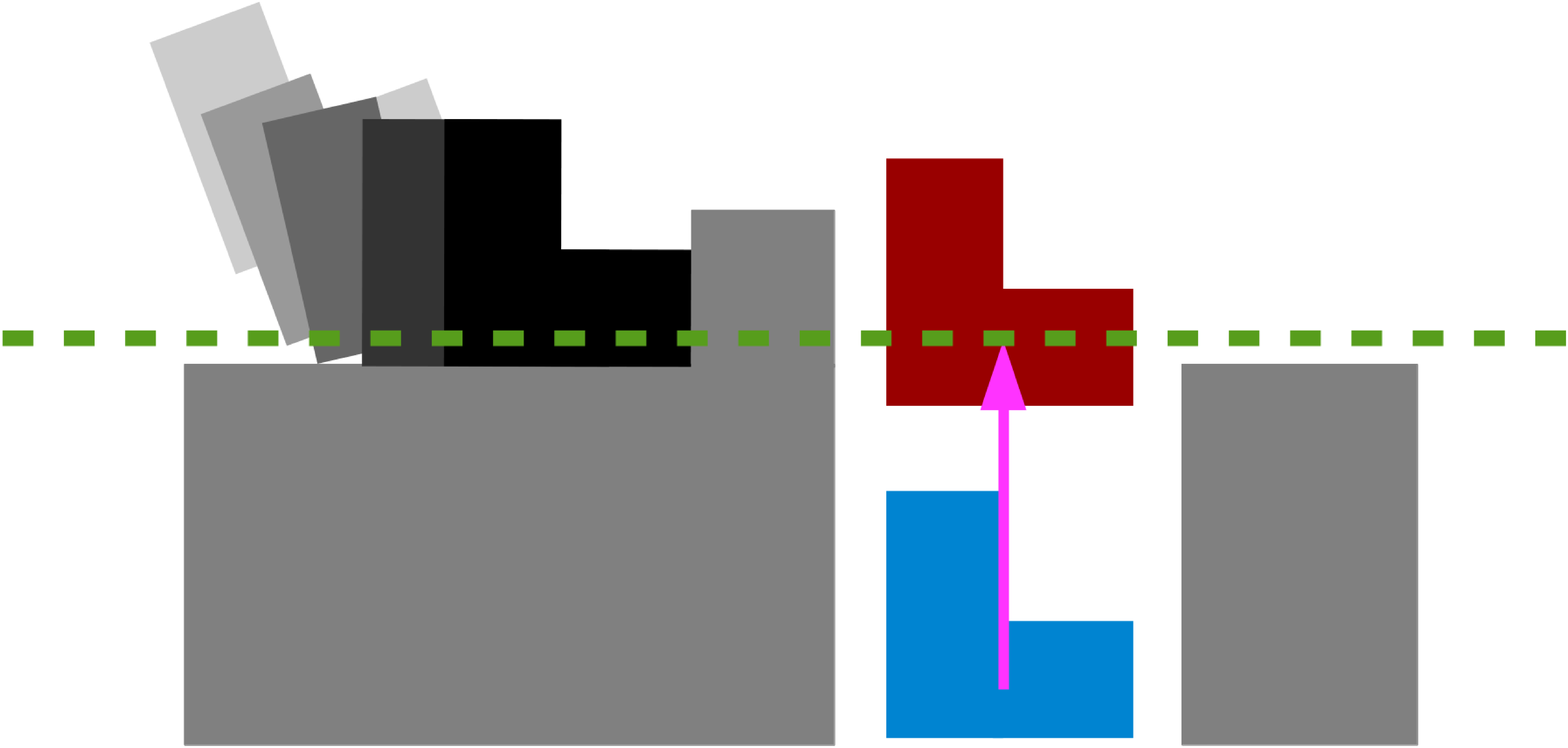}
    }
    \,
    \caption{\small{Our contact motion controller helps mitigate the effects of contact friction. (a) The robot approaches contact while moving towards the goal in blue. (b) The robot makes contact and becomes stuck on the surface, from which we estimate a plane (green) that locally approximates the surface and adjust the goal by $\epsilon_\mathit{adjust}$ shown in magenta to reduce contact force until (c) the robot resumes moving. (d) Alternatively, the robot remains stuck for $i$ iterations until $i \epsilon_\mathit{adjust} = 1$ and the controller terminates.}}
    \label{fig:contactmotioncontroller}
\end{figure*}

We use the Gazebo dynamics simulator in both planar and 3D environments to simulate execution of robot motion including contact with obstacles. In the kinematic simulator used during planning, a PD position controller attempts to reach a target configuration $q_\mathit{target}$ by commanding velocities to the robot. Likewise, we control the simulated robot in Gazebo using a position controller that receives $q_\mathit{target}$  and commands velocities. To safely achieve those velocities in collision and contact, a velocity controller commands forces and torques that move the simulated robot. Unlike the kinematic simulator, which ignores friction and dynamic effects to achieve faster runtime, the dynamic simulator incorporates friction between the robot and the environment. To mitigate the effects of friction in execution, we use a \emph{contact motion controller} illustrated in Figure \ref{fig:contactmotioncontroller} which adjusts $q_\mathit{target}$ to reduce contact forces that cause the robot to become stuck.

When the contact motion controller receives a new target position, it first commands $q_\mathit{target}$ without modification. For the duration of execution $t_\mathit{exec}$, at each iteration the controller records the trajectory of the robot and checks if the robot has become stuck, i.e. if the total motion over a sliding window of the trajectory is below a threshold $\epsilon_\mathit{stuck}$. If the robot is stuck, we assume that the surface on which the robot is stuck can be locally approximated as a plane, which we can estimate from the recent motion of the robot. Once the robot is stuck, the controller then fits a plane defined by point $P_\mathit{plane}$ and normal vector $\overrightarrow{N_\mathit{plane}}$ to the sliding window of the trajectory and projects $q_\mathit{target}$ towards the plane:

\begin{align}
\label{eq:contactmotioncontrol}
q_\mathit{target}' &= q_\mathit{target} + ( \frac{ \overrightarrow{q_\mathit{target}, P_\mathit{plane}} \cdot \overrightarrow{N_\mathit{plane}} }{ \overrightarrow{N_\mathit{plane}} \cdot \overrightarrow{N_\mathit{plane}} } \overrightarrow{N_\mathit{plane}} ) (i \epsilon_\mathit{adjust})
\end{align}

Here, on the $i$th stuck iteration of the controller (i.e. the robot has be stuck for $i$ consecutive iterations of the controller), the controller computes $\overrightarrow{q_\mathit{target},P_\mathit{plane}}$, the vector from $q_\mathit{target}$ to $P_\mathit{plane}$, and projects it onto $\overrightarrow{N_\mathit{plane}}$ to compute an adjustment vector. The target configuration is then moved along the adjustment vector towards the plane by $i \epsilon_\mathit{adjust}$, where $\epsilon_\mathit{adjust}$ is the amount to adjust at each step. If the controller exceeds the time limit $t_\mathit{exec}$ or $i \epsilon_\mathit{adjust} \geq 1$, the controller reports that the robot has become ``completely stuck''. Intuitively, this controller reduces contact forces (and thus the effects of friction) by moving $q_\mathit{target}$ towards the surface of the obstacle approximated by fitting a plane to the trajectory. If the robot continues to move, or if the robot resumes moving after being stuck, the controller commands the original $q_\mathit{target}$. For both $SE(2)$ and $SE(3)$ simulation tests, we used $\epsilon_\mathit{adjust} = 0.01$ (i.e. it will attempt 100 stuck iterations before terminating the motion).

A structurally similar contact motion controller is used with the Baxter robot; however, instead of fitting a plane in $R(7)$ and projecting the target towards it, we use the kinematic simulator to predict the next adjusted target configuration. At a stuck configuration $q_\mathit{stuck}$, we forward-simulate using the kinematic simulator towards $q_\mathit{target}$ for a brief timestep, and record the resulting configuration $q_\mathit{simulated}$. We then interpolate between $q_\mathit{target}$ and $q_\mathit{simulated}$ by $i \epsilon_\mathit{adjust}$ to produce the new adjusted target $q_\mathit{target}'$.
\end{subappendices}

\end{document}